\documentclass[10pt,twocolumn,letterpaper]{article}

\usepackage{cvpr}              

\usepackage{booktabs} 
\usepackage[utf8]{inputenc}
\usepackage{newunicodechar}
\newunicodechar{↔}{\ensuremath{\leftrightarrow}}
\usepackage{wrapfig}
\usepackage{amsmath}
\usepackage{mathtools} 
\usepackage{graphicx}
\usepackage{multirow}
\usepackage{tabularx}
\usepackage{subcaption}
\usepackage{pifont}
\usepackage{amssymb}
\usepackage{float} 
\definecolor{darkgreen}{rgb}{0.0, 0.5, 0.0}  
\definecolor{lightgrey}{rgb}{0.05, 0.05, 0.05}  
\definecolor{myyellow}{rgb}{0.75, 0.75, 0.05}  
\definecolor{cvprblue}{rgb}{0.21,0.49,0.74}
\usepackage[pagebackref,breaklinks,colorlinks,allcolors=cvprblue]{hyperref}

\def\paperID{} 
\def\confName{CVPR}
\def\confYear{2026}

\title{Mitigating Multimodal Hallucinations via Gradient-based Self-Reflection}

\author{Shan Wang$^{1,2,3}$
\and Maying Shen$^{1}$ \and Nadine Chang$^{1}$ 
\and
Chuong Nguyen$^{3}$ \and  Hongdong Li$^{2}$ \and Jose M. Alvarez$^{1}$  
\\
$^{1}$NVIDIA \quad
$^{2}$Australian National University \quad
$^{3}$Data61, CSIRO
}

\begin{document}
\maketitle
\begin{abstract}
Multimodal large language models (MLLMs) achieve strong performance across diverse tasks but remain prone to hallucinations, where outputs are not grounded in visual inputs. This issue can be attributed to two main biases: text–visual bias, the overreliance on prompts and prior outputs, and co-occurrence bias, spurious correlations between frequently paired objects. 
We propose Gradient-based Influence-Aware Constrained Decoding (GACD), an inference-based method, that addresses both biases without auxiliary models, and is readily applicable to existing models without finetuning. The core of our approach is bias estimation, which uses first-order Taylor gradients to understand the contribution of individual tokens—visual features and text tokens—to the current output. Based on this analysis, GACD mitigates hallucinations through two components: (1) suppressing spurious visual features correlated with the output objects, and (2) rebalancing cross-modal contributions by strengthening visual features relative to text. Experiments across multiple benchmarks demonstrate that GACD effectively reduces hallucinations and improves the visual grounding of MLLMs outputs.
\end{abstract}    
\section{Introduction}
\label{sec:intro}
\begin{figure}[ht]
    \centering
    \includegraphics[width=\linewidth]{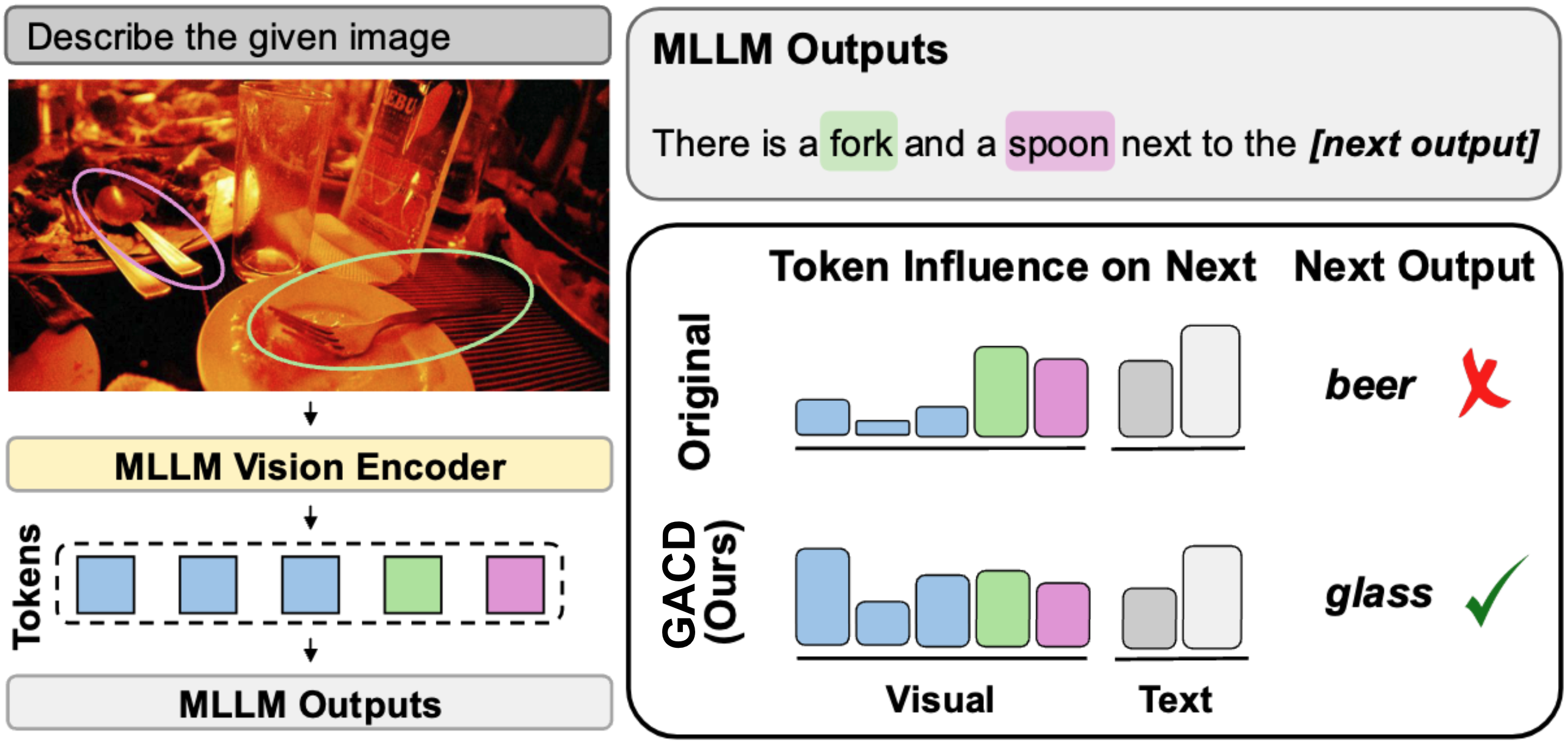}
    \caption{
    Overview of our influence-aware constrained decoding framework, which mitigates hallucinations by regulating token-level influence. It reduces text–visual bias by enhancing visual token influence (\textcolor{blue}{blue} bars) in alignment with the most influential text inputs— prompts (\textcolor{gray}{gray}) or previous outputs (\textcolor{lightgray}{white}). It further mitigates co-occurrence bias through anchor-specific suppression, selectively suppressing visual tokens (\textcolor{darkgreen}{green}, \textcolor{magenta}{magenta}) anchored to previously emitted nouns.
    }
    \label{fig:teaser}
    \vspace{-10pt}
\end{figure}

Recent advances in Multimodal Large Language Models (MLLMs) show strong ability to produce coherent and context-aware content across a wide range of domains~\citep{bai2023qwen, dai2023instructblipgeneralpurposevisionlanguagemodels, chen2024internvl, liu2024llava, ye2024mplug}. Despite their impressive advancements, these models remain prone to hallucination, wherein the generated text is not faithfully grounded in the visual modality~\citep{rohrbach2018object, li2023evaluating}. This limitation poses a critical barrier to establishing trust in the outputs of MLLMs.

The hallucinations observed in MLLMs can be largely attributed to two fundamental biases~\citep{kang2023impact, li2023evaluating, kim2024discovering}. \textbf{Text-visual bias} refers to the excessive reliance on textual information—such as the input prompt and previously generated outputs—while neglecting the visual modality during generation. This bias becomes particularly pronounced in longer sequences, where MLLMs tend to depend more heavily on prior text and increasingly disregard visual cues~\citep{zhou2023analyzing, favero2024multi}. \textbf{Co-occurrence bias} arises from spurious statistical correlations embedded in the training data, which lead models to erroneously predict the presence of non-existent objects based on their frequent co-occurrence with observed objects in the visual inputs~\citep{li2023evaluating}. This bias is particularly challenging to mitigate, and existing approaches largely rely on statistical priors rather than offering statistically agnostic solutions~\citep{kang2023impact,zhou2023solution}.

Existing efforts to mitigate hallucinations in MLLMs can be broadly categorized into inference-based methods, which operate at the decoding stage~\citep{chen2024halc, favero2024multi, leng2024mitigating, park2024convis, woo2024don}, and training-based approaches, which intervene during model optimization~\citep{ben2023mocha, chen2023mitigating, kang2023impact, sun2023aligning, jiang2024hallucination, yue2024less}. Inference-based approaches are valued for their cost-effectiveness, as they avoid the need for additional data collection, data bias examination, or extensive model retraining. However, these methods offer limited insight into the severity of underlying biases, leaving the root causes of hallucination insufficiently understood. In addition, some inference-based methods rely on auxiliary models—such as segmentation networks~~\cite{chen2024halc}, detection systems~~\cite{kan2024catchcomplementaryadaptivetokenlevel} , or even additional MLLMs~\citep{radford2021learning, deng2024seeing, xing2024efuf}—
which introduce extra sources of error, depend on task-specific supervision.

Another limitation of existing methods lies in their lack of granularity when adjusting the underlying biases in MLLMs. Most approaches rely on heuristically tuned priors, which vary across datasets and fail to generalize reliably~\cite{leng2024mitigating, zhao2024mitigating}. Moreover, they apply uniform weighting across all visual features, offering no mechanism to selectively adjust bias at the level of individual features~\cite{zhang2024debiasing,manevich2024mitigatinghallucinationslargevisionlanguage}. This coarse treatment limits their effectiveness in mitigating co-occurrence bias, which arises from spurious statistical correlations between objects that are often represented by distinct visual features.

In this work, we propose an inference-based method that simultaneously addresses both text–visual bias and co-occurrence bias, without relying on auxiliary models or external supervision. 
The core of our approach is the estimation of underlying bias, achieved by quantifying the contribution of individual tokens—both visual features and text tokens—through gradients derived from a first-order Taylor expansion.
Building on this analysis, the method mitigates hallucinations by reweighting tokens via two key components: (1) suppressing the influence of visual features that exhibit strong spurious correlations with the current output token, thereby reducing co-occurrence bias; and (2) rebalancing cross-modal contributions by enhancing the role of visual features to align more closely with that of text tokens in generating the current output. As illustrated in Fig.~\ref{fig:teaser}, our method, GACD, corrects hallucinated predictions—such as the spurious generation of “beer” in the presence of “fork” and “spoon”—by amplifying the contributions of visual tokens unrelated to those nouns, leading to outputs that are more faithfully grounded in the visual modality. Note also that our method is readily applicable to existing MLLMs at inference time.

We summarize our main contributions as follows.
\begin{itemize}
\item We introduce an inference-based method for hallucination mitigation in MLLMs, built on a principled estimation of underlying bias via gradients obtained from a first-order Taylor expansion. This estimation provides a mechanism for understanding and granularly adjusting their influences of individual visual features and text tokens on the generation of the current output token, all without requiring auxiliary models or finetuning.
\item We design two complementary modules: (i) suppression of spurious visual features correlated with the current output token to alleviate co-occurrence bias, and (ii) cross-modal rebalancing to enhance the contributions of visual features relative to text tokens, thereby addressing text–visual bias.
\item 
Extensive experiments demonstrate that GACD mitigates hallucinations and enhances accuracy while maintaining a favorable balance between accuracy and informativeness.
GACD achieves up to $8\%$ increase in overall score on AMBER \cite{wang2023llm}, an $8\%$ F1 boost on POPE \cite{li2023evaluating}, up to $45\%$ improvement in detailness and a $92\%$ accuracy gain on LLaVA-QA90 \cite{liu2024visual}. 
\end{itemize}
\section{Related Work}
\label{sec:relatedwork}

\noindent\textbf{Hallucination and Bias}.
Hallucinations in LLMs often arise from biases in the training data \cite{mckenna2023sources,huang2025survey}, while in MLLMs,
studies \cite{tonmoy2024comprehensive, li2023evaluating, fu2024mmecomprehensiveevaluationbenchmark} show that
hallucinations are closely linked to biases like text-visual and co-occurrence biases.
Additionally, biases related to output position, which increase the risk of hallucination as output length grows, have been examined in \cite{favero2024multi, zhou2023analyzing}. 
Existing methods \cite{li2023evaluating, fu2024mmecomprehensiveevaluationbenchmark, kim2024discovering} typically report only overall statistics, lacking a mathematical, sample-wise bias measurement. This distinction is important, as biases can vary case by case. 
Our approach measures sample-dependent bias via token-level gradient sensitivities, 
revealing how pre-trained MLLM parameters embed these biases ~\cite{kim2019learning,guo2024bias}, and enabling self-reflective hallucination mitigation.

\noindent\textbf{Hallucination Mitigation}. 
Training-related hallucination mitigation methods \cite{chen2023minigpt,jiang2024hallucination,yue2024less,peng2025mitigating,zadeh2025lpoi} are expensive, requiring access to training data and specialized statistical analysis.
Among them, LPOI \cite{zadeh2025lpoi} also employs an object-aware framework, highlighting the effectiveness of modeling object-level information for mitigating hallucinations.
Reinforcement‐learning approaches \cite{xing2024efuf,deng2024seeing,zhai2024hallecontrolcontrollingobjecthallucination} rely on supplementary feedback, often from human annotators or auxiliary LLMs/MLLMs, and the latter may themselves hallucinate.
By contrast, post-decoding techniques modify model logits at inference time without further training or external feedback, making them lightweight add-ons. In text‐only LLMs, such methods aim to align outputs with factual knowledge \cite{chuang2023dola,li2023inference}. In MLLMs, post-decoding strategies emphasize the role of visual inputs \cite{leng2024mitigating,zhao2024mitigating,favero2024multi} and can be classified into image-level and token-level interventions. 
Image-level decoding methods \cite{zhang2024debiasing,manevich2024mitigatinghallucinationslargevisionlanguage} treat all objects in the input image uniformly, limiting their effectiveness in addressing co-occurrence hallucinations.
Existing token-level methods either rely on external segmentation~\cite{chen2024halc} and detection models~\cite{kan2024catchcomplementaryadaptivetokenlevel} 
or lack awareness of object-related decoupling~\cite{woo2024don}.
Moreover, these methods typically introduce an implicit trade-off between accuracy and informativeness, reducing hallucinations at the expense of omitting valid details. 
Attention-based methods~\cite{tang2025intervening,zhang2024redundancy} require careful selection of specific layers and often introduce model-specific adjustments or heuristics. 
In contrast, our GACD directly estimates embedded bias and decouples object-aware visual tokens, enabling sample-specific hallucination mitigation without external data, models, or model-specific adjustments, while achieving a more favorable balance between accuracy and informativeness. 
\section{Method} 
\label{subsec:GACD}
In this section, we provide background on MLLMs, introduce the concept of token influence, and explain how GACD balances token influence to mitigate hallucinations.

\subsection{Background on MLLMs}
MLLMs generate a finite token sequence
\(\mathbf{y}=[y_{1},\dots,y_{M}]\) in response to a visual input (image or video) and a textual prompt.
Let \(\mathcal{V}\) be a finite vocabulary. The prompt is tokenized as
\(\mathbf{t}^{p}=[t^{p}_{1},\dots,t^{p}_{N}]\) with \(t^{p}_{n}\in\mathcal{V}\).
The visual input is encoded by a visual encoder into features, 
which are then projected into the token-embedding space \(\mathbb{R}^{d}\), yielding visual tokens
\(\mathbf{t}^{v}=[t^{v}_{1},\dots,t^{v}_{S}]\) with \(t^{v}_{s}\in\mathbb{R}^{d}\),
where \(d\) is the shared token embedding dimension used for \(\mathcal{V}\).

A MLLM with parameters \(\theta\) computes, at each decoding step \(m\),
a logit vector
\begin{equation}
\label{eq:logits}
\mathbf{z}_m
\;=\;
\pi_{\theta}\!\left(\mathbf{t}^{v},\,\mathbf{t}^{p},\,\mathbf{y}_{<m}\right)
\in \mathbb{R}^{|\mathcal{V}|},
\end{equation}
where
$\mathbf{y}_{<m}=[y_{1},\dots,y_{m-1}] \text{ (empty when } m=1\text{)}$.
This induces a categorical next-token distribution via the softmax
$\sigma:\mathbb{R}^{|\mathcal{V}|}\!\to\!\Delta^{|\mathcal{V}|-1}$:
\begin{equation}
\label{eq:MLLM1}
p_{\theta}\!\left(y_{m}\mid \mathbf{t}^{v},\mathbf{t}^{p},\mathbf{y}_{<m}\right)
\;=\;
\big[\sigma(\mathbf{z}_m)\big]_{y_m}, 
\quad 1\le m\le M,
\end{equation}
where $\sigma(\mathbf{z}_m)\in\Delta^{|\mathcal{V}|-1}$ denotes the probability distribution\footnote{We use “confidence” to denote the model-assigned probability of the emitted token.} over the vocabulary, 
and $[\cdot]_{y_m}$ selects the component corresponding to token $y_m\in\mathcal{V}$.
At inference, $y_m$ is sampled from this categorical distribution (e.g., greedy, beam search).
The sequence likelihood factorizes by the chain rule:
\begin{equation}
\label{eq:MLLM2}
p_{\theta}\!\left(\mathbf{y}\mid \mathbf{t}^{v},\mathbf{t}^{p}\right)
\;=\;
\prod_{m=1}^{M}
p_{\theta}\!\left(y_{m}\mid \mathbf{t}^{v},\mathbf{t}^{p},\mathbf{y}_{<m}\right).
\end{equation}

Given a dataset \(\mathcal{D}\) of \((\mathbf{t}^{v},\mathbf{t}^{p},\mathbf{y})\),
maximum-likelihood training (or fine-tuning) estimates \(\theta^\star\) by maximizing
the conditional log-likelihood.
Pretrained MLLMs encode statistical regularities (including spurious correlations)
from training data in \(\theta^{\star}\); such behavior can be probed without changing
\(\theta^{\star}\) via parameter-dependent analyses (e.g., gradients/attributions or
counterfactual decodings)~\cite{kim2019learning,guo2024bias}, enabling self-reflective
bias interpretation.

\subsection{Gradient-Based Token Influence Estimation}
\label{sec:token_In}
To capture these embeded biases, we interpret how each input token perturbs the output logits.
Let \(\mathbf{z}^{\star}_m \in \mathbb{R}^{|\mathcal{V}|}\) denote the step-\(m\) logits
\(\mathbf{z}^{\star}_m = \pi_{\theta^\star}(\mathbf{t}^{v},\mathbf{t}^{p},\mathbf{y}_{<m})\).
Around a reference sample point \((\mathbf{t}^{v(0)},\mathbf{t}^{p(0)}, \mathbf{y}_{<m}^{(0)})\),
the first-order Taylor expansion~\cite{spivak1980calculus} of the logits \(\mathbf{z}^{\star}_m\) is
\begin{equation}\label{eq:taylor-first}
\begin{aligned}
\mathbf{z}^{\star}_m
\;\approx\; \sum^S_{s=1} \mathbf{g}^v_{ms} {t}^v_s + \sum^N_{n=1} \mathbf{g}^p_{mn} {t}^p_n  + \sum^{m-1}_{i=1} \mathbf{g}^{y}_{mi}  y_i + Const,
\end{aligned}
\end{equation}
where $Const$ denotes other terms that are constant w.r.t., $\mathbf{t}^{v}$ and $\mathbf{t}^{p}$ and the token-wise Jacobians are
\begin{equation}\label{eq:jacobian-notation}
\begin{split}
\mathbf{g}^{v}_{ms}
&\;\coloneqq\;
\left.\frac{\partial\, \mathbf{z}^{\star}_m}{\partial\, \mathbf{t}^{v}_s}\right|_{\mathbf{t}^{v} = \mathbf{t}^{v(0)}},\qquad
\mathbf{g}^{p}_{mn}
\;\coloneqq\;
\left.\frac{\partial\, \mathbf{z}^{\star}_m}{\partial\, \mathbf{t}^{p}_n}\right|_{\mathbf{t}^{p} = \mathbf{t}^{p(0)}},\\
\mathbf{g}^{y}_{mi}
&\;\coloneqq\;
\frac{\partial\, \mathbf{z}^{\star}_m}{\partial\, \mathbf{y}_i}\Big|_{\mathbf{y} =\mathbf{y}_{<m}^{(0)}},
\end{split}
\end{equation}
where \(\big|_{\cdot}\) indicate evaluation at the reference sample point. 
Taylor expansion details are in supplementary Sec.~1. 
Each \(\mathbf{g}^{v}_{ms},\,\mathbf{g}^{p}_{mn},\,\mathbf{g}^{y}_{mi}\)
indicate a small token perturbation in its embedding space to a perturbation of the predict logit vector in \(\mathbb{R}^{|\mathcal{V}|}\).
Following \cite{simonyan2013deep}, we approximate the importance of each input token by the Manhattan norm of its gradient:
\begin{equation}\label{eq:token-influence-emitted}
I^{v}_{ms} \;=\; \|\mathbf{g}^{v}_{ms}\|_1, \quad
I^{p}_{mn} \;=\;\|\mathbf{g}^{p}_{mn}\|_1, \quad
I^{y}_{mi} \;=\; \|\mathbf{g}^{y}_{mi}\|_1,
\end{equation}
and $I^{v}_{ms}[c]$ represents the gradient from the output vocabulary $c$ with respect to each visual tokens.
Aggregating over tokens yields step-\(m\) group-level influences:
\begin{equation}\label{eq:group-influence-emitted}
\texttt{I}^{v}_m \;=\; \sum_{s=1}^{S} I^{v}_{ms},\quad
\texttt{I}^{p}_m \;=\; \sum_{n=1}^{N} I^{p}_{mn},\quad
\texttt{I}^{y}_m \;=\; \sum_{i=1}^{m-1} I^{y}_{mi}.
\end{equation}
These quantities decompose, at the sample level, how visual tokens, prompt tokens, and prior outputs contribute to the logit of \(y_m\), enabling interpretation of bias per sample.

\subsection{Influence-Aware Constrained Decoding}
GACD builds on token influence estimation with two components: (i) Object-aware Visual Token Grouping and (ii) Anchor-specific Influence-weighted Decoding. 
At step \(m\), the former partitions visual tokens into object-related \(\mathbf{t}^{o}\) and unrelated \(\mathbf{t}^{u}\) based on objects detected in \(\mathbf{y}_{<m}\).
The latter extends contrastive decoding~\cite{li2022contrastive} by forming \emph{Anchor-specific} negative guidance logits from pre-mentioned objects and computing a decoding weight \(\alpha_m\) from token-influence measurements.

\begin{figure*}[!th]
    \centering
    \includegraphics[width=0.80\textwidth]{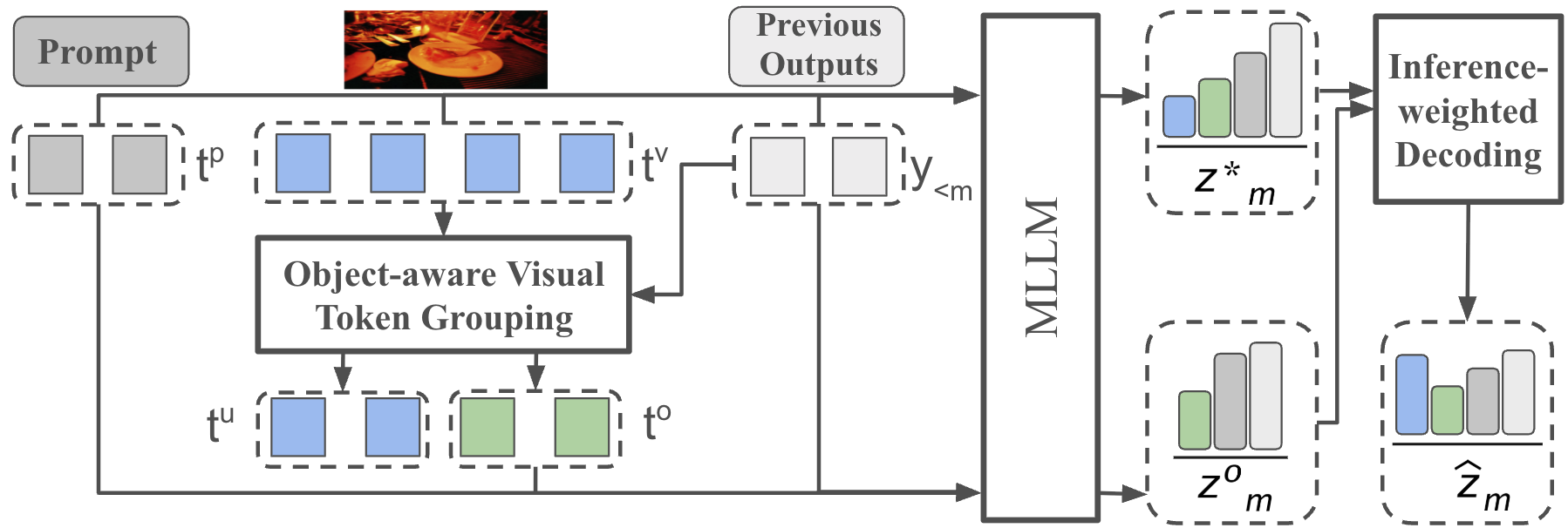}
    \caption{
    Overview of GACD. The method comprises (i) Object-aware Visual Token Grouping and (ii) Anchor-specific Influence-Weighted Decoding. At step $m$, previously mentioned objects are detected from $\mathbf{y}_{<m}$; visual tokens are partitioned into object-related textcolor{darkgreen}{$\mathbf{t}^{o}$} and unrelated \textcolor{blue}{$\mathbf{t}^{u}$} via token influence (Sec.~\ref{sec:token_In}). Anchor-specific Influence-weighted Decoding extends contrastive decoding with token influence, explicitly amplifying the influence of \textcolor{blue}{$\mathbf{t}^u$} to jointly counter text-visual and co-occurrence biases; negative-guidance logits $\mathbf{z}_m^{o}$ are generated from $\{\textcolor{darkgreen}{\mathbf{t}^o}, \mathbf{t}^p, \mathbf{y}_{<m}\}$ to suppress text tokens and anchor-specific visual cues. Grouping is invoked only for noun prediction (where co-occurrence arises between object pairs); for non-noun prediction, we set $\textcolor{darkgreen}{\mathbf{t}^{o}}=\varnothing$ and uniformly amplify all visual tokens to balance text–visual bias.
    } 
    \label{fig:overview}
    \vspace{-0.4cm}
\end{figure*}

\paragraph{Object-aware Visual Token Grouping.}
For each step \(m\), we detect nouns in \(\mathbf{y}_{<m}\) and treat each noun \(y_i\) as an object mention. 
To link a mention to visual evidence, we measure the influence \(I^{v}_{is}\) of visual token \(s\) on step \(i\), 
For every noun \(y_i\), the visual token with maximal influence is selected to form a mask \(\mathcal{M}_{is}\).
The cumulative object mask at step \(m\) aggregates all prior noun-linked tokens:
\begin{equation}\label{eq:marsk}
\mathcal{M}_{ms} = \mathbf{1}\!\left[ \sum_{i=1}^{m-1} \mathcal{M}_{is} > 0 \right], 
\end{equation}
where $\mathcal{M}_{is} = \mathbf{1}\!\left[\, y_i \in \text{Noun} \;\wedge\; s = \arg\max_{j} I^{v}_{ij} \,\right]$ and \(\mathbf{1}[\cdot]\) is the indicator and \(\wedge\) is logical AND.

The mask \(\mathcal{M}_{ms}\) identifies visual tokens linked to nouns emitted before \(m\). 
We then partition the visual tokens into \emph{object-related ($\mathbf{t}^{o}$)} and \emph{unrelated-to-objects ($\mathbf{t}^{u}$)} sets via a Hadamard product:
\begin{equation}\label{eq:token-split}
\mathbf{t}^{o} \;=\; \mathbf{t}^{v} \odot \mathcal{M}_{m}, 
\quad
\mathbf{t}^{u} \;=\; \mathbf{t}^{v} \odot ( \mathbf{1} - \mathcal{M}_{m} ).
\end{equation}
Object-related and unrelated influences at step \(m\) are
\begin{equation}\label{eq:influence-split}
\texttt{I}^{o}_m \;=\; \sum_{s=1}^{S} \|\mathbf{g}^{v}_{ms}\|_{1}\,\mathcal{M}_{ms},
\;
\texttt{I}^{u}_m \;=\; \sum_{s=1}^{S} \|\mathbf{g}^{v}_{ms}\|_{1}\,(1-\mathcal{M}_{ms}).
\end{equation}
Masking and grouping are applied only during noun prediction (mitigate co-occurrence hallucination from object pairs). 
For non-noun steps, all elements in \(\mathcal{M}_m\) are set to \(0\), yielding an empty \(\mathbf{t}^{o}\).

\paragraph{Anchor-specific Influence-weighted Decoding.}
Let  \(\mathbf{z}^{o}_m=\pi_{\theta^\star}(\mathbf{t}^{o},\mathbf{t}^{p},\mathbf{y}_{<m})\)
the \emph{anchor-specific} negative logits and \(\mathbf{z}^{\star}_m=\pi_{\theta^\star}(\mathbf{t}^{v},\mathbf{t}^{p},\mathbf{y}_{<m})\)
be the \emph{original} logits. We adjust logits by
\begin{equation}\label{eq:cd}
\hat{\mathbf{z}}_m
\;=\; (1+\alpha_m)\,\mathbf{z}^{\star}_m \;-\; \alpha_m\,\mathbf{z}^{o}_m,
\end{equation}
with \(\alpha_m\ge 0\). 
In the probability space, moving along
\(\mathbf{z}^{\star}_m-\mathbf{z}^{o}_m\) increases the KL divergence
\(D_{\mathrm{KL}}(\sigma(\mathbf{z}^{\star}_m)\|\sigma(\mathbf{z}^{o}_m))\) (see Sec.~2 in supplementary). 
The original logits distribution $\mathbf{z}^{\star}_m$ can be viewed as $\pi_{\theta^*}(\mathbf{t}^u , \mathbf{t}^o,\mathbf{t}^p, \mathbf{y}_{<m})$, i.e., a joint distribution that additionally depends on $\mathbf{t}^u$ compare to $\mathbf{z}^{o}_m$. 
Increasing the KL divergence therefore emphasizes the contribution of tokens $\mathbf{t}^u$, which are unrelated to previous mentioned objects, thereby mitigating co-occurrence bias in noun prediction. For non-noun steps, $\mathbf{t}^u$ coincides with $\mathbf{t}^v$, meaning that all visual tokens are emphasized. This adjustment helps reduce text–visual hallucination.

When analyzing token influence of $\hat{\mathbf{z}}_m$ in Eq.~\ref{eq:cd}, the chain rule shows that $\mathbf{t}^u$ occur only in the original logits $\mathbf{z}^{\star}_m$ and are amplified by $(1 + \alpha_m)$, whereas other inputs ($\mathbf{t}^o$, $\mathbf{t}^p$, $\mathbf{y}_{<m}$) also contribute to $\mathbf{z}^o_m$ and therefore undergo smaller influence changes.
Let \(\tilde{\texttt{I}}^{o}_m,\tilde{\texttt{I}}^{p}_m,\tilde{\texttt{I}}^{y}_m\)
denote group-level influences computed on the negative branch \(\mathbf{z}^{o}_m\)
(analogous to~\eqref{eq:group-influence-emitted}).
We then choose $\alpha_m$ so that the influence of $\mathbf{t}^u$ matches the \emph{dominant text} level, \(\texttt{I}^{t}_m \coloneqq \max(\texttt{I}^{p}_m,\texttt{I}^{y}_m)\). 
Aligning $\mathbf{t}^u$ influence with the question prompt $\texttt{I}^p_m$ is crucial for visually grounded responses, while balancing with previous outputs $\texttt{I}^y_m$ prevents visual forgetting.
\begin{equation}
\begin{aligned}
    \alpha_m = 
    \frac{\texttt{I}^{t}_m-\texttt{I}^v_m}{\texttt{I}^v_m-\tilde{\texttt{I}_m}^o +\tilde{\texttt{I}}^{t}_m -\texttt{I}^{t}_m },
    \tilde{\texttt{I}}^{t}_m =
    \begin{cases} 
    \tilde{\texttt{I}}^{p}_m & \text{if } \texttt{I}^p_m \geq \texttt{I}^y_m\\
    \tilde{\texttt{I}}^{y}_m & \text{otherwise} 
     \end{cases}
    \end{aligned}
\label{eq:alpha2}
\end{equation}

Unlike existing decoding methods \cite{zhou2023analyzing,favero2024multi,leng2024mitigating}, which rely on adaptive plausibility constraints (e.g., prediction confidence) and require experimental tuning to determine optimal thresholds, 
our approach explicitly enforces non-negativity on the influence of object-related visual and prompt tokens. This corresponds to the following upper-bound condition:
\begin{equation}\label{eq:constraint}
0 \;\le\; \alpha_m \;\le\; 
\min\!\left\{
\frac{\texttt{I}^{o}_m}{\,\tilde{\texttt{I}}^{o}_m-\texttt{I}^{o}_m\,},\;
\frac{\texttt{I}^{p}_m}{\,\tilde{\texttt{I}}^{p}_m-\texttt{I}^{p}_m\,}
\right\}.
\end{equation}

\paragraph{Sample-dependent early stopping.}
Additionally, since hallucinations are more likely in long generations \cite{zhou2023analyzing,peng2025mitigating}, we introduce a sample-dependent stopping criterion based on visual influence. Specifically, if the visual influence ratio $r^v_m$ of the token following the end-of-sequence (EOS) falls below a threshold $\epsilon$,
\begin{equation}\label{eq:es}
r^{v}_m \;\coloneqq\; 
\frac{\texttt{I}^{v}_m}{\texttt{I}^{v}_m+\texttt{I}^{p}_m+\texttt{I}^{y}_m}
< \epsilon
\quad\text{and}\quad
y_{m-1}=\text{EOS}.
\end{equation}
Early stopping is triggered to prevent further output generation with minimal visual grounding.

\section{Experiments}

\label{sec:experiments}

The proposed method is evaluated for both the open-ended generative tasks and the discriminative tasks. We 
use Amber~\cite{wang2023llm}, MSCOCO~\cite{lin2014microsoft} and LLaVa-QA90~\cite{liu2024visual} datasets for 
the generative task, and Amber~\cite{wang2023llm} and POPE~\cite{li2023evaluating} datasets on the discriminative tasks. 

\subsection{Evaluation Metrics}
For generative image captioning, we focus on object hallucination and follow \cite{deng2024seeing}
report the Caption Hallucination Assessment with Image Relevance (CHAIR) \cite{rohrbach2018object} score, which includes sentence-level ($hal$,$C_S$) and instance-level ($cha$,$C_I$) percentages, instance-level recall ($R$,$cov$), and the average generated length ($Len$)~\footnote{
Since shorter outputs can trivially lower CHAIR scores at the expense of informativeness.}, as well as co-occurrence object hallucination ($cog$) and the overall $score$ as suggested by \cite{wang2023llm}.
For generative VQA, follow \cite{leng2024mitigating,huang2024opera} GPT-4V \cite{achiam2023gpt} is used to score both accuracy (Acc) and detailedness (Det) on a scale of 10.
For discriminative tasks, hallucination manifests as a `yes/no' misclassification we report both accuracy and F$1$ score.

\subsection{Implementation Details}
The maximum output length is set to 256 across all models, with other model parameters kept at their defaults. 
Gradients are obtained via PyTorch’s torch.autograd.grad on the input tokens.
Noun tokens are identified using the spaCy library via its \texttt{en\_core\_web\_sm} model.
To prevent excessive modifications, we set the maximum
$\alpha_m$ to 5 for discriminative tasks and 3 for generative tasks.
We empirically set the early stopping thresholds $\epsilon$ as follows: LLaVA-v1.5 and LLaVA-v1.6: $7\%$, InstrucBLIP: $25\%$, mPLUG-Owl2: $2.5\%$, and InternVL2: $10\%$.
All experiments are performed on an NVIDIA A40 GPU with batch size of 1. Unless otherwise specified, we use greedy sampling~\cite{graves2013generating}.

\subsection{Results on Open-ended Generation}
In this section, we compare against SOTA alignment-based method RLAIF-V and contrastive decoding methods VCD, M3ID, and AVISC,
on the AMBER and MSCOCO datasets, as presented in Tab.~\ref{tab:AMBER}, Tab.~\ref{tab:coco_chair}. Additionally, we evaluate our method against VCD on the LLaVA-QA90 dataset, presented in Tab.~\ref{tab:LLaVA_QA90}.
Our method outperforms most existing approaches across various baseline models and datasets, highlighting its robustness and reliable performance across different data types and model architectures. 
Specifically, we surpass image-level contrastive decoding methods like VCD and M3ID, demonstrating its effectiveness in operating at the token level and adapting to individual samples. 
Furthermore, compared to the token-level AVISC, our method excels, likely due to its object awareness and adaptability to fluctuating bias levels.
The results further demonstrate that our method effectively mitigates hallucinations while preserving information. 

\begin{table}[htbp]
\scriptsize
\centering
\caption{Results on the AMBER Dataset.} 
\vspace{-8pt}
\begin{tabularx}{0.48\textwidth}{p{16pt} p{25pt} c*{9}{X}}
\toprule
\multicolumn{2}{c}{\multirow{3}{*}{Method}} & 
\multicolumn{4}{c}{Generative Task} & 
\multicolumn{4}{c}{Discriminative Task} & 
\multirow{2}{*}{\rotatebox{90}{Score}}\\
\cmidrule(lr){3-6} \cmidrule(lr){7-10} 
& & cha & cov  & hal  & cog & acc & P  & R  & F1 &  \\
& & $\downarrow$ & $\uparrow$ & $\downarrow$ & $\downarrow$ & $\uparrow$ &  $\uparrow$ & $\uparrow$ & $\uparrow$ & $\uparrow$ \\
\midrule
\multirow{6}{*}{\shortstack{LLaVA\\v1.5}} 
& base & 7.8 & \textbf{51.0} & 36.4 & 4.2 & 72.0 & \textbf{93.2} & 62.4 & 74.7 & 83.5\\
& RLAIFv~\citenum{yu2024rlaif}  & 6.6 & 49.7 & 32.0 & 2.9 & 76.7 & 92.0 & 78.1 & 84.5 & 89.0\\
& VCD~\citenum{leng2024mitigating} & 6.7 & 46.4 & 32.6 & 3.5 & 71.3 & 91.1 & 62.3 & 74.3 &  83.8\\
& M3ID~\citenum{favero2024multi} & 6.2 & 50.5 & 29.3 & 2.8 & 72.4 & 91.8 & 64.1 & 75.5 & 84.7 \\
& AVISC~\citenum{woo2024don} & 6.5 & 50.2 & 34.8 & 2.7 & 73.8 & 89.7 & 68.4 & 77.6 & 85.5 \\
& Ours & \textbf{5.6} & \textbf{51.0} & \textbf{24.3} & \textbf{1.8} & \textbf{80.3} & 82.9 & \textbf{89.3} & \textbf{86.0} & \textbf{90.2}\\
\midrule
\multirow{6}{*}{\shortstack{Instruct\\BLIP}} 
& base & 8.8 & \textbf{52.2} & 38.2 & 4.4 & 76.5 & 84.5 & 79.0 & 81.7 & 86.5 \\
& RLAIFv~\citenum{yu2024rlaif} & 7.6 & 47.7 & 29.9 & 2.8 & 76.5 & 84.5 & 79.0 & 81.7 & 87.1 \\
& VCD~\citenum{leng2024mitigating} & 7.9 & 49.7 & 36.7 & 3.7 & 75.9 & 83.5 & 79.3 & 81.3 & 86.7 \\
& M3ID~\citenum{favero2024multi} & 7.3 & 49.2 & 33.8 & 3.7 & 75.8 & 84.4 & 77.9 & 81.0 & 86.9\\
& AVISC~\citenum{woo2024don} & 7.1 & 48.8 & 34.4 & 4.3 & 75.9 & 83.4& \textbf{79.5} & 81.4 & 87.2 \\
& Ours & \textbf{6.0} & 49.4 & \textbf{26.6} & \textbf{2.4} & \textbf{78.1} & \textbf{88.8} &  76.6 & \textbf{82.2} & \textbf{88.1}\\
\midrule
\multirow{6}{*}{\shortstack{mPLUG\\Owl2}} & base & 10.6 & 52.0 & 39.9 & 4.5 & 75.6 & \textbf{95.0} & 66.9 & 78.5 & 84.0\\
& RLAIFv~\citenum{yu2024rlaif} & 7.8 & 50.5 & 35.7 & 4.1 & 81.2 & 90.8 & 79.7 & 84.9 & 88.6\\
& VCD~\citenum{leng2024mitigating} & 8.0 & 51.3 & 35.3 & 4.1 & 75.6 & 83.5 & 78.8 & 81.1 & 86.6 \\
& M3ID~\citenum{favero2024multi} & 7.8 & 51.7 & 34.9 & 4.1 & 75.9 & 83.5 & 79.3 & 81.3 & 86.8 \\
& AVISC~\citenum{woo2024don} & 10.9 & 50.5 & 35.5 & 4.4 & 82.1 & 90.7 & 81.4 & 85.8 & 87.5 \\
& Ours & \textbf{7.5} & \textbf{53.6} & \textbf{34.7} & \textbf{4.0} & \textbf{82.1} & 87.0 & \textbf{86.2} & \textbf{86.6} & \textbf{89.6}\\
\midrule
\multirow{6}{*}{\shortstack{LLaVA\\v1.6}} 
& base     & 9.9 & 56.7 & 47.4 & 4.3  & 80.3 & 82.9 & \textbf{89.3} & 86.0 & 88.5 \\
& RLAIFv  & 9.0 & 53.6 & 46.1 & 3.42 & 80.8 & 83.9 & 88.9 & 86.3 & 88.6 \\
& VCD      & 9.5 & 52.7 & 46.3 & 3.78 & 79.9 & 83.1 & 87.6 & 85.4 & 88.0 \\
& M3ID     & 9.2 & 50.1 & 45.3 & 3.3  & 80.4 & 83.2 & 88.8 & 85.9 & 88.4 \\
& AVISC    & 9.2 & 50.7 & 47.5 & 3.2  & 80.6 & 83.5 & 88.2 & 85.8 & 88.3 \\
& Ours     & \textbf{8.7} & \textbf{58.3} & \textbf{43.8} & \textbf{2.5} & \textbf{81.2} & \textbf{85.2} & 88.8 & \textbf{87.0} & \textbf{89.2} \\
\midrule
\multirow{6}{*}{\shortstack{Qwen2\\VL}} 
& base     & 6.4 & 70.4 & 54.8 & 5.9  & 82.9 & \textbf{91.6} & 82.2 & 86.6 & 90.1 \\
& RLAIFv  &  5.8 & 69.4 & 54.1 & 5.5 & 83.5 & 91.2 & 82.6 & 86.7 & 90.4\\
& VCD      & 6.5 & 69.1 & 53.7 & 5.3 & 82.7 & 90.9 & 82.3 & 86.4 & 90.0\\
& M3ID     & 6.3 & 68.8 & 53.5 & 5.1 & 83.0 & 91.0 & 82.8 & 86.7 & 90.2\\
& AVISC    &  6.3 & 69.0 & 53.9 & 5.0 & 82.8 & 91.1 & 82.5 & 86.6& 90.1\\
& Ours     & \textbf{4.9} & \textbf{71.8} & \textbf{44.7} & \textbf{3.7} & \textbf{84.4} & 88.1 & \textbf{89.2} & \textbf{87.1} & \textbf{91.1} \\
\midrule
\multirow{6}{*}{\shortstack{Intern\\VL2}} 
& base     & 8.1 & 69.6 & 59.0 & 5.2 & 84.0 & 87.3 & \textbf{88.8} & 88.0 & 90.0 \\
& RLAIFv  & 8.0 & 68.4 & 59.3 & 4.9 & 84.2 & 87.7 & 88.5 & 88.1 & 90.1 \\
& VCD      & 8.5 & 68.7 & 58.6 & 5.0 & 82.9 & 87.0 & 88.4 & 87.7 & 89.6 \\
& M3ID     & 8.4 & 69.2 & 58.9 & 5.4 & 83.7 & 86.8 & 88.4 & 87.6 & 89.6 \\
& AVISC    & 8.4 & 68.9 & 59.1 & 4.8 & 84.0 & 87.7 & 86.8 & 87.2 & 89.4 \\
& Ours     & \textbf{7.9} & \textbf{69.8} & \textbf{57.8} & \textbf{3.7} & \textbf{84.7} & \textbf{88.2} & \textbf{88.8} & \textbf{88.5} & \textbf{90.3} \\
\bottomrule
\end{tabularx}
\label{tab:AMBER}
\vspace{-0.3cm}
\end{table}

\begin{table}[htbp]
\centering
\caption{Open-ended Generation Results Using the CHAIR Metric on the MSCOCO Subset Following \cite{deng2024seeing}.}
\vspace{-0.2cm}
\scriptsize 
\begin{tabularx}{0.48\textwidth}{l | l | *{6}{X}}
\toprule
Models & Metrics & Baseline & VCD & M3ID & AVISC & Ours \\
\midrule
\multirow{3}{*}{LLaVA-v1.5} & $C_S \downarrow$ & 48.8 & 44.8 & 44.5& 46.4 & \textbf{41.0} \\
 & $C_I \downarrow$ & 13.4 & 12.8& 12.1 & 13.4 &\textbf{10.9}  \\
 & $R \uparrow$ & \textbf{78.6} & 76.8  & 77.0 & 76.3  & 77.3  \\
 & $Len \uparrow$ & 99.8 & 89.8 &85.1 & 90.5 & 85.0  \\
 \midrule
\multirow{3}{*}{InstructBLIP} & $C_S \downarrow$ & 57.8 & 63.4 & 57.3 &58.9& \textbf{47.4}  \\
 & $C_I \downarrow$ & 16.5 & 19.6 & 16.1 & 17.8 & \textbf{13.4}  \\
 & $R \uparrow$ & \textbf{73.6} & 71.2 & 72.5 & 70.6 & 72.3 \\
 & $Len \uparrow$ & 101.3 & 95.5 & 100.1 & 99.6 & 93.9   \\
 \midrule
 \multirow{3}{*}{mPLUG-Owl2} & $C_S \downarrow$ & 59.2 & 52.7 & 52.4  & 58.3 & \textbf{45.0} \\
 & $C_I \downarrow$ & 17.6 & 16.1 & 15.8 & 17.5  & \textbf{12.4} \\
 & $R \uparrow$ & \textbf{75.8}  & 73.2 & 72.7 &75.6  & 74.9 \\
 & $Len \uparrow$ & 105.3 & 93.6 & 92.6 & 99.5 & 83.5   \\
\bottomrule
\end{tabularx}
\label{tab:coco_chair}
\vspace{-0.4cm}
\end{table}

\noindent\textbf{Hallucination Mitigation}.
Our approach reduces hallucination by up to $33\%$ at sentence-level ($hal$ in Tab.~\ref{tab:AMBER} and $C_S$ in Tab.~\ref{tab:coco_chair}) and $32\%$ at instance-level ($cha$ in Tab.~\ref{tab:AMBER} and $C_I$ in Tab.~\ref{tab:coco_chair}), demonstrating its effectiveness in mitigating overall hallucinations.
It also effectively mitigates co-occurrence hallucinations, with reductions of up to $57\%$ for $cog$ in Tab.~\ref{tab:AMBER}. 
Accuracy gains of up to $92\%$ (Tab.~\ref{tab:LLaVA_QA90}) further demonstrate that our model improves alignment with the input image, highlighting its ability to jointly address text–visual and co-occurrence biases (Sec.~\ref{sec:ablation}).

\noindent\textbf{Information Preservation}.
Our method also enhances information preservation, with recall ($cov$ in Tab.~\ref{tab:AMBER} and $R$ in Tab.~\ref{tab:coco_chair}) dropping by an average of only $1.1\%$, compared to an average drop of $3.2\%$ in other methods, and even increasing by $3.1\%$ when using the baseline mPLUG-Owl2 on the AMBER dataset.
Higher recall indicates that our model retrieves a broader range of objects from visual inputs. Additionally, results in Tab.~\ref{tab:LLaVA_QA90} an increase of up to $45\%$ in detailedness ($Det$), further demonstrating our method's effectiveness in retrieving all relevant visual details and mitigating visual forgetting.
\begin{table}[htbp]
\centering
\scriptsize
\vspace{-0.3cm}
\caption{Results on LLaVA-QA90, settings following \cite{leng2024mitigating}. }
\vspace{-8pt}
\begin{tabular}{lcccccc}
\toprule
\multirow{2}{*}{\parbox{0.7cm}{\vspace{3pt} Method}} & \multicolumn{2}{c}{LLaVA-v1.5} & \multicolumn{2}{c}{IntructBLIP} & \multicolumn{2}{c}{mPLUG-Owl2} \\
\cmidrule(lr){2-3} \cmidrule(lr){4-5} \cmidrule(lr){6-7}
& Acc$\uparrow$ & Det$\uparrow$ & Acc$\uparrow$ & Det$\uparrow$ & Acc$\uparrow$ & Det$\uparrow$  \\
\midrule
base & 3.23 & 3.54 & 3.84 & 4.07 & 4.07 & 4.33 \\ 
VCD  & 4.15 & 3.85 & 4.23 & 4.69 & 4.52 & 4.64 \\ 
M3ID & 4.57 &3.96 & 4.67 & 4.61 & 4.44 & 5.18 \\
AVISC & 4.88 & 3.87 & 4.32 & 4.27 & 4.75 & 5.12 \\
RLAIF-V & 5.79 & 4.74 & 5.27 & 4.62 &5.03 & 5.33\\
Ours & \textbf{6.20} & \textbf{5.13} & \textbf{6.28} & \textbf{4.77} &  \textbf{6.69} &  \textbf{6.28} \\
\bottomrule
\end{tabular}
\label{tab:LLaVA_QA90}
\vspace{-10pt}
\end{table}


\subsection{Results on Discriminative Task}

We next evaluate our method on discriminative tasks using AMBER (discriminative VQA) and POPE (existence VQA), with results shown in Tab.~\ref{tab:AMBER} and Tab.~\ref{tab:POPE}. 
Our approach achieves a better balance between precision and recall, yielding consistently higher F1 scores and improved accuracy. Notably, unlike competing methods that degrade Intern-VL2, ours preserves its performance via bias awareness.
Category-wise analysis further shows heterogeneous gains, with improvements varying across question types.

\begin{table}
\scriptsize
\centering
\vspace{-10pt}
\caption{Results on POPE in MSCOCO Adversarial Setting.}
\vspace{-10pt}
\begin{tabularx}{0.48\textwidth}{l *{9}{X}}
\toprule
\multirow{3}{*}{Method} & \multicolumn{2}{c}{LLaVA} & \multicolumn{2}{c}{Instruct} & \multicolumn{2}{c}{mPLUG} &\multicolumn{2}{c}{Intern} \\
 & \multicolumn{2}{c}{v1.5} & \multicolumn{2}{c}{BLIP} & \multicolumn{2}{c}{Owl2} &\multicolumn{2}{c}{VL2} \\
\cmidrule(lr){2-3} \cmidrule(lr){4-5} \cmidrule(lr){6-7} \cmidrule(lr){8-9} 
& Acc $\uparrow$ & F1 $\uparrow$ & Acc $\uparrow$ & F1 $\uparrow$ & Acc $\uparrow$ & F1 $\uparrow$ & Acc $\uparrow$ & F1 $\uparrow$ \\
\midrule
base & 80.9 & 81.6 & 79.8 & 81.4 & 72.5 & 77.5 &\textbf{85.8} & \textbf{85.0}\\ 
VCD & 80.9 & 81.3 & 79.6 & 79.5 & 74.2 & 78.8 & 83.2 & 82.2\\
M3ID & 81.7 & 81.8 & 81.0 & 81.6 & 75.6 & 79.1 & 83.5 & 82.1 \\ 
AVISC & 81.2 & 81.6 & 81.8 & 81.9 & 80.9 & 79.7 & 85.3 & 84.6 \\ 
Woodpecker & 80.5 & 80.6 & 79.0 & 78.6 & 77.5 & 76.9 & 85.7 & 84.8\\ 
Ours & \textbf{83.5} & \textbf{82.1} & \textbf{82.5} & \textbf{82.1} & \textbf{84.2}  & \textbf{83.7} & \textbf{85.8}  & \textbf{85.0}\\
\bottomrule
\end{tabularx}
\label{tab:POPE}
\end{table}

\begin{figure}[ht]
    \centering
        \begin{subfigure}[b]{0.225\textwidth}
            \includegraphics[width=\textwidth]{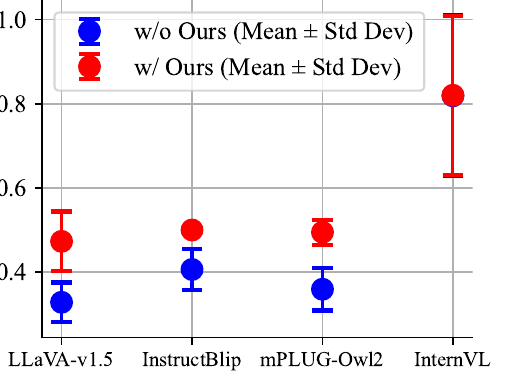}
            \caption{Visual Influence Ratio}
            \label{fig:pope_ratio}
        \end{subfigure}
        \begin{subfigure}[b]{0.245\textwidth}
            \includegraphics[width=\textwidth]{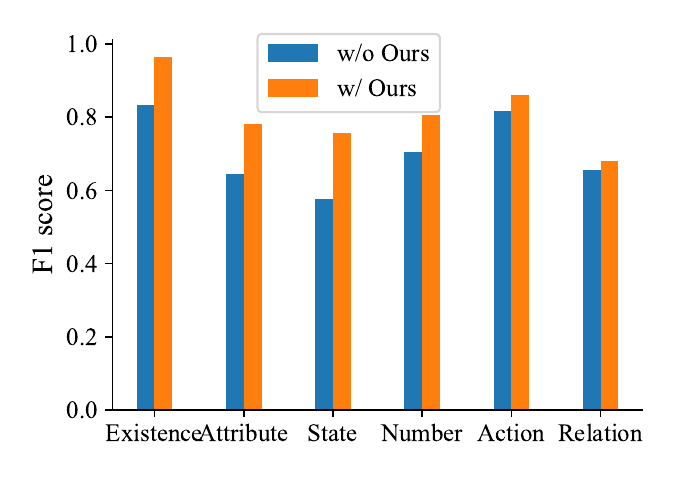}
            \caption{F1 across Question Categories}
            \label{fig:amber_d}
        \end{subfigure}
        \vspace{-0.5cm}
        \caption{(a) Visual influence ratios across the POPE dataset, illustrating variation across MLLMs. Our method successfully increases the visual influence ratio when it falls below  $50\%$. (b) F1 scores for the AMBER discriminative task using LLaVA-v1.5 are consistently improved by our method, with particularly notable gains in the existence and state categories.}
        \label{fig:pope}
        \vspace{-0.4cm}
\end{figure}

\noindent\textbf{Variation in Improvement Across Question Categories}.
Fig.~\ref{fig:amber_d} presents F1 scores across various question categories using LLaVA-v1.5 \cite{liu2024llava}. Our method improves performance across all categories, with the largest gains in existence, attributes, and state—categories strongly tied to visual cues, benefiting from enhanced visual token influence.

\noindent\textbf{Variation in Improvement Across MLLMs}.
Our method achieves the most significant improvement on mPLUG-Owl2 (Tab.~\ref{tab:POPE}) and on LLaVA-v1.5  (Tab.~\ref{tab:AMBER}), 
Consistent performance gains are observed across modern MLLMs (LLaVA-v1.6, Qwen2-VL, and InternVL2). Compared with static heuristics (e.g., VCD), which may degrade performance due to overcorrection, our method maintains stable improvements across models.
We further analyze and find that performance variation is correlated with the baseline visual influence ratio.
Fig.~\ref{fig:pope_ratio} 
presents the visual influence ratios in object existence VQA, showing that LLaVA-v1.5 exhibits the lowest visual contribution, followed by mPLUG-Owl2. This lower baseline visual influence ratio allows our method to make more impactful adjustments. In contrast, InternVL2 has an original visual influence ratio exceeding $50\%$, resulting in minimal improvement when our method is applied.
The strong performance of InternVL2 can be attributed to its original high visual influence ratio, further validating the motivation behind our approach.

\begin{figure*}[!htbp]
    \centering
    \includegraphics[width=0.90\textwidth]{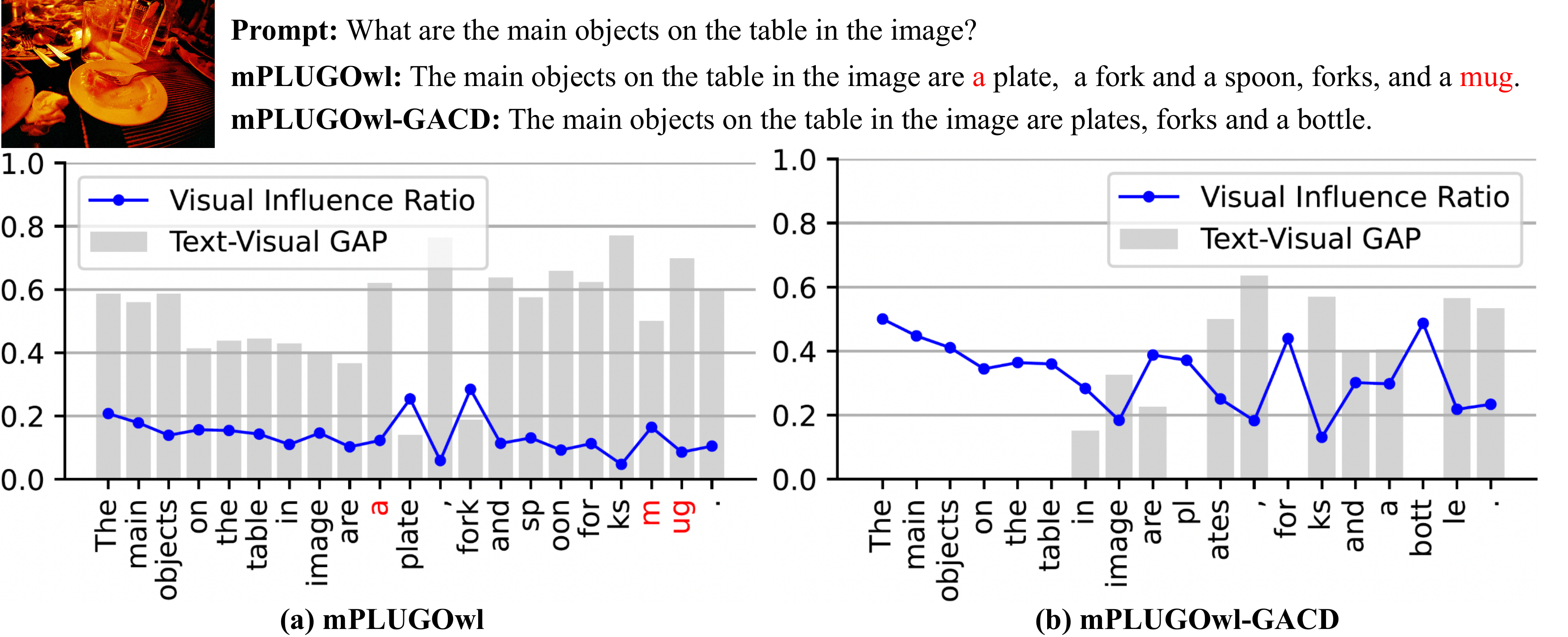}
    \vspace{-6pt}
    \caption{Comparison of visual influence ratios $r^v_m$ and Text-Visual GAP, with and without our GACD. (a) Without GACD, mPLUG-Owl2 shows a low initial \textcolor{blue}{visual influence ratio}, 
    punctuation marks and suffixes naturally have low visual influence, while objects start with higher influence that declines as the sequence grows. \textcolor{red}{Hallucinations} tend to occur when the visual ratio is low. The text-visual gap confirms that text dominates the influence on predictions. (b) With GACD, the visual influence ratio increases overall and mitigates the decrease over the sequence length. The text tokens only domain influence in predictions less related to the visual, reducing hallucination.}
    \label{fig:cap_llava}
    \vspace{-0.4cm}
\end{figure*}

\section{Ablation Study} 
\label{sec:ablation}

In this section, we first analyze text–visual and co‐occurrence biases, then evaluate the contributions of our proposed components. We also detail the gradient-computation methods and norm selection. Additional ablation studies and hyperparameter settings are provided in the Supplementary Material.
\noindent\textbf{Text-Visual Bias Analysis}. 
 Fig.~\ref{fig:pope_ratio} shows that with the exception of InternVL2, MLLMs (LLaVA-v1.5, InstructBLIP, and mPLUG-Owl2) rely more on text prompt than on visual input. Likely due to MLLMs' training process, this tendency is common in MLLMs, where multimodal features are aligned with language tokens after extensive text-based pre-training, causing language components to dominate decision-making. 
 GACD effectively increases overall visual influence to match that of object-present question prompts in POPE (Fig.~\ref{fig:pope_ratio}).  
 In the open-ended generation task, we further observe the visual influence ratio $r^v_m$ and the Text-Visual GAP, defined as $\max(\max(r^p_m, r^y_m) - r^v_m,0)$, the difference between the text influence ratio and the visual influence ratio when text influence is dominant \footnote{$r^p_m$ and $r^y_m$ are derived in the same manner as $r^v_m$ in \eqref{eq:es}.}.
 Observations in Fig.~\ref{fig:cap_llava} 
 also highlight the text-dominant influence typical of MLLMs.
GACD counteracts this by boosting the influence of visual tokens when aligning them with prompts and previous outputs, leading to higher prediction confidence and a reduction in hallucinations (Fig.~1
in supplementary).
Additionally, the nature of the output token affects the visual influence ratio. For instance, punctuation marks or suffixes tend to have a lower visual influence ratio. This is intuitive, as these tokens rely more on linguistic context and are less dependent on visual information. This observation highlights the value of our GACD framework delivering sample-dependent, token-specific hallucination mitigation.

\begin{figure*}[!htbp]
    \centering
    \begin{minipage}{\textwidth}
    \begin{subfigure}[t]{0.33\textwidth}
        \includegraphics[width=\textwidth]{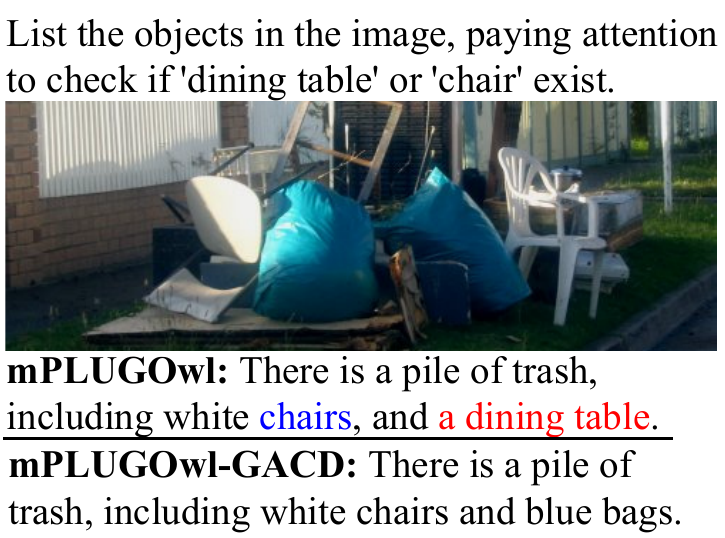}
        \vspace{-12pt}
        \caption{Co-occurrence Hallucination}
        \label{fig:cr_example}
    \end{subfigure}
    \hfill
    \begin{subfigure}[t]{0.38\textwidth}
        \includegraphics[width=\textwidth]{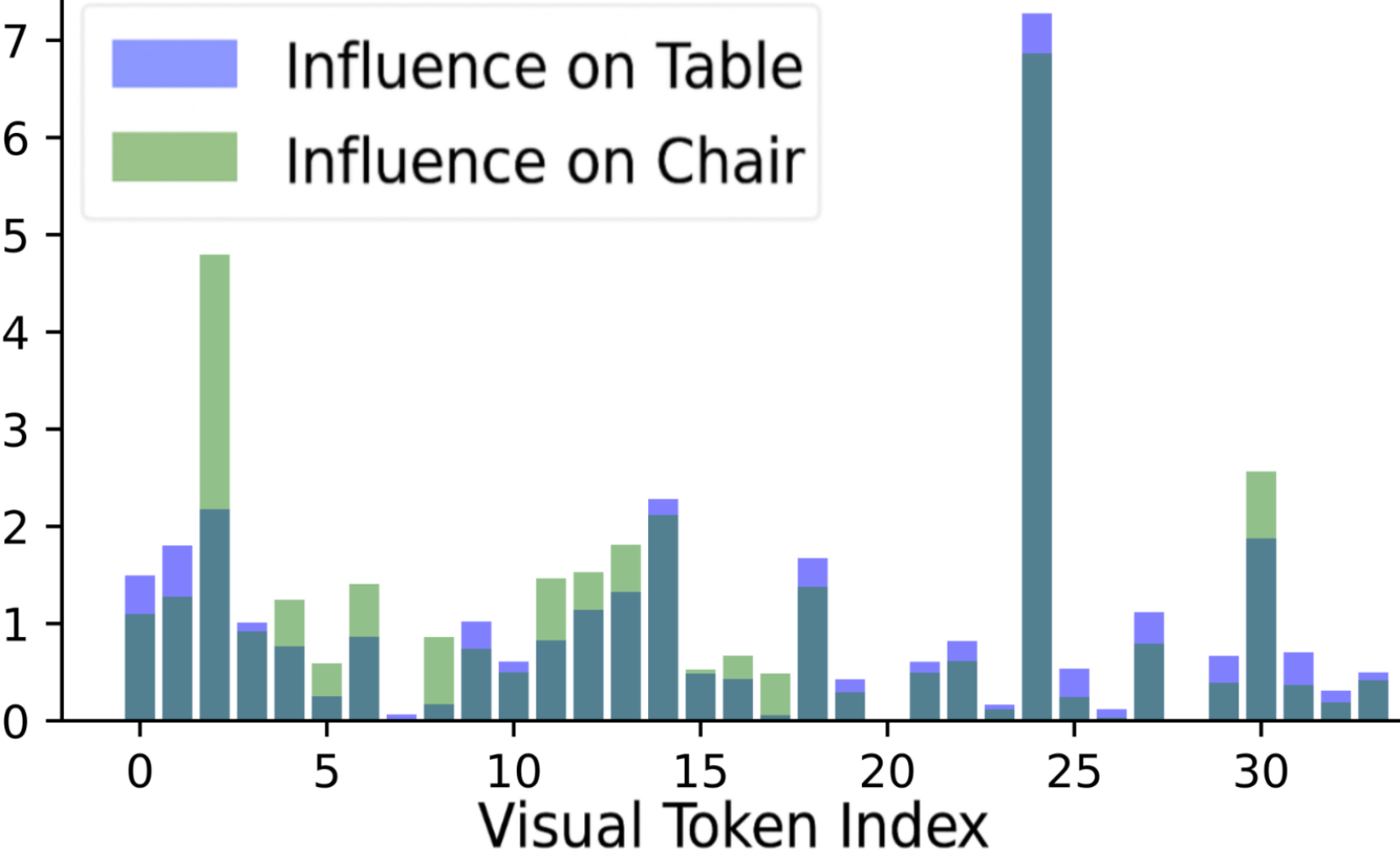}
        \vspace{-12pt}
        \caption{Individual Visual Token Influence}
        \label{fig:cr_vis}
    \end{subfigure}
    \hfill
    \begin{subfigure}[t]{0.27\textwidth}
        \includegraphics[width=\textwidth]{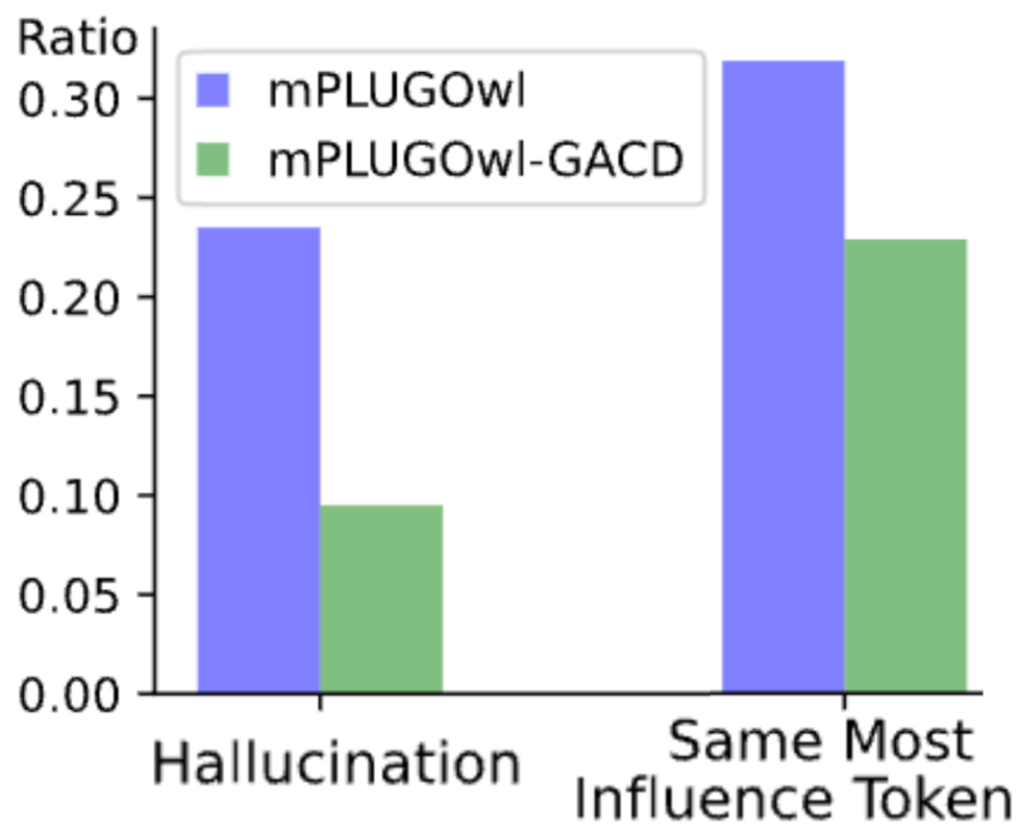}
        \vspace{-12pt}
        \caption{Overall Statistics}
        \label{fig:cr_100}
    \end{subfigure}
    \end{minipage}
    \vspace{-6pt}
    \caption{Co-occurrence hallucination of `table' in the presence of `chair'. (a) 
    Comparison of outputs with and without GACD. (b) Visualization of individual visual token influence indicates that the visual token with index 24, which has the highest influence on the hallucinated `table', also holds the highest influence on `chair'. (c) Summary statistics for 100 chair-only and 100 table-only images, showing the hallucination rate and the percentage of cases where both objects share the same most influential visual token (as illustrated in b). GACD effectively reduces both metrics.}
    \label{fig:noun_reduction}
    \vspace{-15pt}
\end{figure*}

\noindent\textbf{Co-occurrence Bias Analysis}. 
Fig.~\ref{fig:cr_example} shows an example where mPLUG-Owl2 incorrectly predicts `dining table' due to the presence of a `chair'.
In Fig.~\ref{fig:cr_vis}, the influence of individual visual tokens on hallucinated prediction $I_{ms}(\text{`table'})$ and $I_{ms}(\text{`chair'})$ shows that they share the same most influential visual token: $s=24$. 
These visualizations indicate that the influence distribution over tokens is typically sharply peaked, so selecting the single most influential token gives a clean and interpretable attribution signal while reducing noise from low-influence tokens.
We further collected 100 chair-only and 100 table-only images from MSCOCO evaluation dataset \cite{lin2014microsoft}.

Results in Fig.~\ref{fig:cr_100} show that when only a `chair' or `table' exists in the image, the other object is hallucinated in $23.5\%$ of cases, with $31.9\%$ sharing the same most influential token,
indicating that the `Same Most Influential Token' phenomenon is common in co-occurrence hallucinations. Our GACD effectively reduces such hallucination where both `table' or `chair' are predicted in single-object images.

\begin{table}[htbp]
\centering
\scriptsize
\vspace{-0.2cm}
\caption{Component Analysis Using the CHAIR Metric.}
\vspace{-0.2cm}
\begin{tabularx}{0.48\textwidth}{l l | *{4}{X} }
\toprule
\multirow{3}{*}{Models} & VA &  & \checkmark & \checkmark & \checkmark \\
 & CR &  &  & \checkmark & \checkmark \\
 & ES  &  & & & \checkmark \\
\midrule
\multirow{4}{*}{LLaVA-v1.5} & $C_S \downarrow$ & 48.8 & 46.4 & 46.2 & \textbf{41.0} \\
 & $C_I \downarrow$ & 13.4 & 11.6 & 11.3 & \textbf{10.9}  \\
 & $R \uparrow$ & 78.6 & 79.0 &  \textbf{79.4} & 77.3  \\
 & $Len \uparrow$ & \textbf{99.8} &  95.6 & 95.5 &  85.0  \\
 \midrule
\multirow{4}{*}{InstructBLIP} & $C_S \downarrow$ & 57.8 & 53.6 & 53.2 & \textbf{47.4}  \\
 & $C_I \downarrow$ & 16.5 & 15.1 & 14.0 & \textbf{13.4}  \\
 & $R \uparrow$ & 73.6 & \textbf{75.3} & 74.6 & 72.3 \\
 & $Len \uparrow$ & 101.3 & \textbf{108.4} &105.7 & 93.9   \\
 \midrule
 \multirow{4}{*}{mPLUG-Owl2} & $C_S \downarrow$ & 59.2  & 52.6 &52.3 & \textbf{45.0} \\
 & $C_I \downarrow$ & 17.6 &14.4  & 14.2 & \textbf{12.4} \\
 & $R \uparrow$ & 75.8 & \textbf{78.2}& 78.0 & 74.9 \\
 & $Len \uparrow$ & \textbf{105.3} & 95.6 & 95.5 & 83.5   \\
\bottomrule
\end{tabularx}
\label{tab:ablation}
\vspace{-0.3cm}
\end{table}

\noindent\textbf{Component Analysis}. 
To assess the effectiveness of each component in our proposed method, we conducted the following variants: 1) Visual Amplification (VA) only: 
visual amplification is applied to all visual tokens ($\mathbf{t}^v$), including during noun predictions.
2) Co-occurrence Hallucination Reduction (CR): object-related visual tokens are detected, and $\mathbf{t}^u$ is amplified during noun predictions.
3) Our full model, with early stopping (ES).
Tab.~\ref{tab:ablation} demonstrates that each component of our method contributes to the overall performance.
VA significantly reduces hallucinations while improving object recall.
CR further mitigates co-occurrence bias, a residual form of the text-visual bias addressed by VA, resulting in additional hallucination reduction.
Both VA and CR achieve these gains without introducing trade-offs.
When necessary, the ES mechanism shortens outputs to 
effectively reduce hallucinations, with only a slight recall trade-off.

\noindent\textbf{Gradient Computation}.
\label{sec:gradient} 
Our method obtains gradients directly through PyTorch’s `torch.autograd.grad' on input tokens, eliminating the need for manual derivations and enabling straightforward reproducibility.
For comparison, we evaluate Integrated Gradients (IG) \cite{enguehard2023sequential, lundstrom2022rigorous,kapishnikov2021guided} using the SIG  \cite{enguehard2023sequential} implementation; results for this ablation on the POPE MSCOCO adversarial setting are shown in Tab.~\ref{tab:study_ig}. 
`IG' denotes the SIG-based results, while “Ours” refers to our direct gradient method. 
Both achieve comparable accuracy and F1‐score, but the direct-gradient variant is substantially more efficient.

\begin{table}[!htbp]
\centering
\scriptsize 
\vspace{-0.2cm}
\caption{Gradient Methods on the POPE MSCOCO Dataset(Adv.)}
\vspace{-0.2cm}
\begin{tabularx}{0.48\textwidth}{p{1.5cm}*{4}{X}}
\toprule
\multicolumn{2}{c}{Methods} &  Accuracy & F1 & Speed(ms) \\
\midrule
\multirow{2}{*}{MPLUG-Owl2} & IG \cite{enguehard2023sequential} & 83.4 & 82.9 & 20335 \\
& Ours  & 84.2 & 83.7 & 385 \\
\bottomrule
\end{tabularx}
\label{tab:study_ig}
\vspace{-0.2cm}
\end{table}

\noindent\textbf{Norm Analysis}.
We also study the effect of norm selection on token influence, comparing L1 (Manhattan), L2 (Euclidean), and L$\infty$ (infinity) norms. 
The L1 norm emphasizes individual token contributions, the L2 norm reflects overall influence, and the L$\infty$ norm focuses on the strongest activation. As shown in Tab.~\ref{tab:norm}, the L1 norm yields the best performance, supporting the intuition that it effectively captures influence magnitude across tokens and channels. 

\begin{table}[htbp]
\centering
\vspace{-0.2cm}
\scriptsize 
\caption{Norm Strategies on the POPE MSCOCO Dataset(Adv.)}
\vspace{-0.2cm}
\begin{tabularx}{0.48\textwidth}{*{9}{X}}
\toprule
\multirow{2}{*}{Norm} & \multicolumn{2}{c}{LLaVA-v1.5} & \multicolumn{2}{c}{InstructBLIP} & \multicolumn{2}{c}{mPLUG-Owl2} \\
\cmidrule(lr){2-3} \cmidrule(lr){4-5} \cmidrule(lr){6-7} 
& Acc & F1  & Acc & F1 & Acc & F1 \\
\midrule
L1 & \textbf{83.5} & \textbf{82.1} & \textbf{82.5} & \textbf{82.1} & \textbf{84.2}  & \textbf{83.7} \\
L2 &  83.2&  81.9 &  79.5 & 79.6  &  83.2  & 82.9  \\
L$_{\infty}$ &  83.4&  82.0 &  82.1  & 81.8  &  80.8  &  80.6 \\
\bottomrule
\end{tabularx}
\label{tab:norm}
\vspace{-0.3cm}
\end{table}

\noindent\textbf{Cost and Runtime}.
We further analyze computational cost and runtime. The visual encoder is executed only once, and the second pass operates on a small set of tokens. 
On the POPE dataset, this yields a $101.44\%$ increase in average,
comparable to decoding-based methods (e.g., VCD) that also require additional guidance computation. 

\begin{table}[h]
\centering
\scriptsize 
\caption{Runtime Comparison on POPE MSCOCO Dataset(Adv.)}
\label{tab:tflops}
\begin{tabularx}{0.48\textwidth}{*{4}{X}}
\toprule
LLaVA-v1.5      & TFLOPs & Runtime (ms) &  Increase \\
\midrule
base   & $9.68$ & 191  & -- \\
VCD        & $19.37$  & 383 & $+100.10\%$ \\
Ours        & $19.49$  & 385  & $+101.44\%$ \\
\bottomrule
\end{tabularx}
\vspace{-0.3cm}
\end{table}


\section{Conclusion}
\label{sec:conclusion}
In conclusion, we introduce a gradient-based self-reflection method to estimate token influence and quantitatively estimate bias severity. This estimation enables the identification of object-related visual tokens, which are then integrated into an influence-aware constrained decoding framework. This framework effectively mitigates both text-visual and co-occurrence biases, reducing hallucinations. Our method operates without requiring additional resources such as costly fine-tuning, extra models, or data statistics.
Furthermore, to reduce text-visual bias in long-generated sequences, we propose a sample-dependent stopping criterion based on the proposed visual influence.

\noindent\textbf{Limitations}.
Our method is limited to white-box MLLMs, as it requires access to gradients. 
Its effectiveness depends on the baseline MLLM's original visual influence ratio, and the importance of visual information.
For instance, existence questions are primarily rely on visual, whereas relational questions are less direct and require inference beyond visual.
As a post-processing technique, our method does not involve model training. In future work, we aim to explore how insights from GACD can guide and improve training strategies for enhanced visual perception in MLLMs.

\section*{Acknowledgments}
This work was conducted during an internship at NVIDIA. Hongdong Li is also partially supported by an ARC Discovery Grant DP220100800.

{
    \small
    \bibliographystyle{ieeenat_fullname}
    \bibliography{main}
}

\clearpage
\setcounter{page}{1}
\setcounter{figure}{0}
\setcounter{table}{0}
\setcounter{section}{0}
\maketitlesupplementary

\section{First Order Taylor Expansion}
\label{sec:Taylor}
Let \(\mathbf{z}_m^{\star} \in \mathbb{R}^{|\mathcal{V}|}\) denote the step-\(m\) logits
\(\mathbf{z}_m^{\star} = \pi_{\theta^\star}(\mathbf{t}^{v},\mathbf{t}^{p},\mathbf{y}_{<m})\).
Around a reference point \((\mathbf{t}^{v(0)},\mathbf{t}^{p(0)}, \mathbf{y}_{<m}^{(0)})\),
the detailed first-order Taylor expansion of the logit \(\mathbf{z}_m^{\star}\) is
\begin{equation}\label{eq:taylor-first-d}
\begin{aligned}
\mathbf{z}_m^{\star}
\;\approx\; &
\underbrace{\mathbf{z}_m^{\star(0)}}_{\displaystyle
\pi_{\theta^\star}(\mathbf{t}^{v(0)},\mathbf{t}^{p(0)},\mathbf{y}^{(0)}_{<m})}
\;+\;
\sum_{s=1}^{S}\, \mathbf{g}^{v}_{ms}\!\left(\mathbf{t}^{v}_s - \mathbf{t}^{v(0)}_s\right) \\
& \;+\;
\sum_{n=1}^{N}\, \mathbf{g}^{p}_{mn}\!\left(\mathbf{t}^{p}_n - \mathbf{t}^{p(0)}_n\right)
\;+\;
\sum_{i=1}^{m-1}\, \mathbf{g}^{y}_{mi}\!\left(\mathbf{y}_i - \mathbf{y}^{(0)}_i\right) \\
= & \sum^S_{s=1} \mathbf{g}^v_{ms} {t}^v_s + \sum^N_{n=1} \mathbf{g}^p_{mn} {t}^p_n  + \sum^{m-1}_{i=1} \mathbf{g}^{y}_{mi}  y_i \\
& + \underbrace{ \mathbf{z}_m^{\star(0)} - \sum^S_{s=1} \mathbf{g}^v_{ms} {t}^{v(0)}_s - \sum^N_{n=1} \mathbf{g}^p_{mn} {t}^{p(0)}_n  - \sum^{m-1}_{i=1} \mathbf{g}^{y}_{mi}  y^{(0)}_i}_{Const} , \\
= & \sum^S_{s=1} \mathbf{g}^v_{ms} {t}^v_s + \sum^N_{n=1} \mathbf{g}^p_{mn} {t}^p_n  + \sum^{m-1}_{i=1} \mathbf{g}^{y}_{mi}  y_i + Const,
\end{aligned}
\end{equation}
where the (token-wise) Jacobians are
\begin{equation}\label{eq:jacobian-notation}
\mathbf{g}^{v}_{ms}
\;\coloneqq\;
\left.\frac{\partial\, \mathbf{z}^{\star}_m}{\partial\, \mathbf{t}^{v}_s}\right|_{\mathbf{t}^{v} = \mathbf{t}^{v(0)}},\qquad
\mathbf{g}^{p}_{mn}
\;\coloneqq\;
\left.\frac{\partial\, \mathbf{z}^{\star}_m}{\partial\, \mathbf{t}^{p}_n}\right|_{\mathbf{t}^{p} = \mathbf{t}^{p(0)}},\qquad
\mathbf{g}^{y}_{mi}
\;\coloneqq\;
\frac{\partial\, \mathbf{z}^{\star}_m}{\partial\, \mathbf{y}_i}\Big|_{\mathbf{y} =\mathbf{y}_{<m}^{(0)}},
\end{equation}
and \(\mathbf{z}_m^{\star(0)} =
\pi_{\theta^\star}(\mathbf{t}^{v(0)},\mathbf{t}^{p(0)},\mathbf{y}^{(0)}_{<m})\).
Here \(\big|_{\cdot}\) denotes evaluation at the reference point.
Each \(\mathbf{g}^{v}_{ms},\,\mathbf{g}^{p}_{mn},\,\mathbf{g}^{y}_{mi}\)
maps a small token perturbation in its corresponding embedding space to a perturbation of the
logit vector in \(\mathbb{R}^{|\mathcal{V}|}\).
And $Const$ denotes all other terms that are constant w.r.t., the $\mathbf{t}^{v}, \mathbf{t}^{p}$.


\section{Interpreting Contrastive Decoding through KL Divergence}
\label{sec:KL}

Kullback-Leibler (KL) divergence can be used to interpret contrastive decoding, It measures the divergence between the reference distribution $p_{\theta^{\star}}(y_{cm}|  \mathbf{t}^{o}, \mathbf{t}^{p}, y_{<m})$ to the $\mathbf{t}^{u}$ joint distribution $p_{\theta^{\star}}(y_{cm}| \mathbf{t}^{v}, \mathbf{t}^{p}, y_{<m})$, where $\mathbf{t}^{v} = \{\mathbf{t}^{u}, \mathbf{t}^{o}\}$.
\begin{equation}
\begin{aligned}
\mathrm{D}_{KL} & = \sum_c
p_{\theta^{\star}}(y_{cm}| \mathbf{t}^{v}, \mathbf{t}^{p}, y_{<m})\text{log}\big( \frac{p_{\theta^{\star}}(y_{cm}| \mathbf{t}^{v}, \mathbf{t}^{p}, y_{<m})}{p_{\theta^{\star}}(y_{cm}| \mathbf{t}^{o},  \mathbf{t}^{p}, y_{<m})} \big) \\
& =
\sum_c p_{\theta^{\star}}(y_{cm}| \mathbf{t}^{v}, \mathbf{t}^{p}, y_{<m})(\text{log}(p_{\theta^{\star}}(y_{cm}| \mathbf{t}^{v}, \mathbf{t}^{p}, y_{<m})) - \text{log}(p_{\theta^{\star}}(y_{cm}| \mathbf{t}^{o}, \mathbf{t}^{p}, y_{<m}))) \\
& =
\sum_c p_{\theta^{\star}}(y_{cm}| \mathbf{t}^{v}, \mathbf{t}^{p}, y_{<m})([\pi_{\theta^*}(\mathbf{t}^v,\mathbf{t}^p)_{m}]_{c} - \text{log}(\sum\exp(\pi_{\theta^*}(\mathbf{t}^v,\mathbf{t}^p)_{m})) \\ 
& - [\pi_{\theta^*}(\mathbf{t}^{o},\mathbf{t}^p)_{m}]_{c} + \text{log}(\sum\exp(\pi_{\theta^*}(\mathbf{t}^{o}, \mathbf{t}^p)_{m}))) \\
& =
\sum_c p_{\theta^{\star}}(y_{cm}| \mathbf{t}^{v}, \mathbf{t}^{p}, y_{<m})(\underbrace{[\pi_{\theta^*}(\mathbf{t}^v,\mathbf{t}^p)_{m} -  \pi_{\theta^*}(\mathbf{t}^o,\mathbf{t}^p)_{m}}_{\text{adjustment term}}]_{c} + Const),
\end{aligned}
\end{equation} 
where $p_{\theta^{\star}}(y_{cm}| \mathbf{t}^{v}, \mathbf{t}^{p}, y_{<m}) = \mathbf{\sigma}(\pi_{\theta^*}(\mathbf{t}^v,\mathbf{t}^p)_{m})$, $p_{\theta^{\star}}(y_{cm}| \mathbf{t}^{o}, \mathbf{t}^{p}, y_{<m}) = \mathbf{\sigma}(\pi_{\theta^*}(\mathbf{t}^{o},\mathbf{t}^p)_{m})$ and $c$ represents a class in the predefined vocabulary. The adjustment term increases the KL divergence, thereby emphasizing the impact of visual tokens. 

\section{Derivation of $\alpha$ Computation}
\label{sec:alpha}
We choose $\alpha_m$ such that the influence of $\mathbf{t}^u$ matches the dominant text level. 
For clarity, we present the derivation for the case where the prompt is dominant; the case where the previous output tokens are dominant is analogous and leads to the same derivative up to a straightforward substitution. Given $\hat{\texttt{I}}^v_m = (1+\alpha_m) \texttt{I}^v_m - \alpha_m \tilde{\texttt{I}}^o_m $ and $\hat{\texttt{I}}^p_m = (1+\alpha_m) \texttt{I}^p_m - \alpha_m \tilde{\texttt{I}}^p_m $, we enforce $\hat{\texttt{I}}^v_m = \hat{\texttt{I}}^p_m$, and solve for $\alpha_m$:
\begin{equation}
\begin{aligned}
(1+\alpha_m) \texttt{I}^v_m - \alpha_m \tilde{\texttt{I}}^o_m &= (1+\alpha_m) \texttt{I}^p_m - \alpha_m \tilde{\texttt{I}}^p_m \\
\alpha_m (\texttt{I}^v_m-\tilde{\texttt{I}_m}^o +\tilde{\texttt{I}}^{p}_m -\texttt{I}^{p}_m) &= \texttt{I}^p_m - \texttt{I}^v_m \\
\alpha_m &= 
    \frac{\texttt{I}^{p}_m-\texttt{I}^v_m}{\texttt{I}^v_m-\tilde{\texttt{I}_m}^o +\tilde{\texttt{I}}^{p}_m -\texttt{I}^{p}_m }.
\end{aligned}
\end{equation}

\section{MLLMs Architectures}
\label{sec:model_detail}
Tab.~\ref{tab:mllm_detail} shows detailed information about the vision encoder and LLM components of the MLLMs used in our experiments.
\begin{table}[H] 
\centering
\vspace{-0.4cm}
\caption{Details of the used MLLM architectures.}
\footnotesize
\vspace{-0.2cm}
\begin{tabularx}{\textwidth}{l*{3}{X}}
\toprule
MLLMs & Vision encoder & LLM  \\
\midrule
LLaVA-v1.5 (7B) & CLIP-L-336px & Vicuna-v1.5-7B \\
LLaVA-v1.5-13B & CLIP-L-336px & Vicuna-v1.5-13B \\
LLaVA-v1.6 & CLIP-L-336px & Vicuna-v1.5-7B \\
InstructBLIP (7B) & BLIP-2 & Vicuna-v1.1-7B \\
InstructBLIP-13B & BLIP-2 & Vicuna-v1.1-13B \\
mPLUG-Owl2 & CLIP-L & LLaMA-2-7B \\
InternVL2-4B & InternViT-300M-448px & Phi-3-mini-128k-instruct \\
Qwen2-VL-7B & QwenViT & Qwen2-7B \\
\bottomrule
\end{tabularx}
\vspace{-0.3cm}
\label{tab:mllm_detail}
\end{table}

\section{Results on MMBench}
\label{sec:MMBench}
We further evaluate our method on MMBench \cite{liu2025mmbench}. The results in Tab.~\ref{tab:MMbench} indicate that our method improves the overall performance and achieves consistent improvements across MLLMs on Coarse Perception (CP). 
This outcome aligns with the intended effect of our method, as its focus on increasing visual influence is directly linked to improving coarse perception capabilities. 
For other metrics, our method yields minor improvements due to the possible reason that 
certain abilities, such as Logical Reasoning (LR), rely more on the language component of MLLMs and cannot be enhanced solely by increasing visual influence.
\begin{table}[htbp]
\centering
\vspace{-0.2cm}
\caption{Results on MMBench Dataset.}
\footnotesize
\vspace{-0.2cm}
\begin{tabularx}{\textwidth}{c*{9}{X}}
\toprule
MLLMs & Method &
Overall & CP & FP-S & FP-C & AR & LR & RR\\
\midrule
\multirow{2}{*}{LLaVA-V1.5} & base & \textbf{62.3} & 68.5 & \textbf{69.6} & \textbf{57.7} & 73.1 & \textbf{29.9} & 54.7 \\
& ours & 61.8 & \textbf{73.2} &62.6 &53.0&\textbf{73.3}&27.8&\textbf{57.8}\\
\midrule
\multirow{2}{*}{mPLUG-Owl2} & base & 63.5 & 68.1 & \textbf{69.1} & \textbf{55.8}& \textbf{78.4} &37.0 &57.0 \\
& ours &\textbf{65.0} &\textbf{72.6} &66.6 &53.0 &76.0 &\textbf{41.6} &\textbf{63.0}\\
\bottomrule
\end{tabularx}
\label{tab:MMbench}
\vspace{-0.5cm}
\end{table}

\section{Results on MM-Vet}
\label{sec:MMVet}

\begin{table}[htbp]
\centering
\vspace{-0.5cm}
\caption{Results on MM-Vet dataset.}
\vspace{-0.2cm}
\footnotesize
\begin{tabularx}{\textwidth}{c*{9}{X}}
\toprule
MLLMs & Method &
Rec & OCR & Know & Gen & Spat & Math & Total \\
\midrule
\multirow{2}{*}{LLaVA-V1.5} & base & 32.9 & 20.1 & \textbf{19.0} & 20.1 & \textbf{25.6 }& 5.2 & 28.0 \\
& ours & \textbf{38.9} & 2\textbf{4.9 }& 15.0  & 15.5& 24.9& \textbf{7.7} & \textbf{28.9}\\
\midrule
\multirow{2}{*}{InstructBlip} & base & 32.4 & 14.6 & 16.5 & \textbf{18.2} & \textbf{18.6} & 7\textbf{.7} & 26.2 \\
& ours & \textbf{40.5} & \textbf{18.0} & \textbf{18.7} & 17.4& 14.9& 3.8 & \textbf{26.6}\\
\midrule
\multirow{2}{*}{mPLUG-Owl2} & base & 36.1 &19.4&\textbf{29.8}&19.4&23.9&\textbf{7.7}&27.3 \\
& ours & \textbf{45.0} &\textbf{26.4} &27.9&\textbf{25.9} &\textbf{24.8} & 3.8&\textbf{33.9}\\
\bottomrule
\end{tabularx}
\label{tab:MMVet}
\vspace{-0.4cm}
\end{table}

The evaluation on MM-Vet \cite{yu2023mm} in Tab.~\ref{tab:MMVet} shows that our method achieves consistent overall (Total) improvement, along with enhancements in recognition (Rec) and Optical Character Recognition (OCR), indicating its effectiveness in improving visual recognition. However, its performance varies across other metrics, including knowledge (Know), generalization (Gen), spatial awareness (Spat), and math (Math), suggesting that our method, which focuses on token influence balancing, may not effectively enhance the generalization ability of MLLMs.

\section{Results on ScienceQA and Vizwiz}
We evaluate our method on two complementary multimodal benchmarks. ScienceQA~\cite{lu2022learn} integrates images, textual context, and curriculum knowledge, requiring models to perform structured multimodal reasoning. VizWiz~\cite{gurari2018vizwiz}, in contrast, consists of visual questions collected from blind users and features real-world challenges such as low-quality images, conversational queries, and unanswerable cases. These datasets jointly assess both reasoning under structured multimodal contexts and robustness in unconstrained real-world settings.
As shown in Table~\ref{tab:scienceqa_vizwiz}, our approach consistently improves over the LLaVA-1.5 baseline. These gains demonstrate the effectiveness of our hallucination mitigation strategy in enhancing visual grounding across both knowledge-driven and real-world VQA tasks.

\label{sec:ScienceQA}
\begin{table}[htbp]
\centering
\caption{Comparison of LLaVA-1.5 and our method on ScienceQA and VizWiz datasets.}
\vspace{-0.3cm}
\footnotesize
\begin{tabular}{l l c c}
\hline
MLLMs & Method & ScienceQA(\%) $\uparrow$ & VizWiz(\%) $\uparrow$ \\
\hline
\multirow{2}{*}{LLaVA-V1.5} & base  & 66.2 & 48.7 \\
          & Ours  & \textbf{68.7} & \textbf{52.8} \\
\hline
\end{tabular}
\label{tab:scienceqa_vizwiz}
\vspace{-0.5cm}
\end{table}

\section{Results on MME}

Our evaluation on MME \cite{fu2024mmecomprehensiveevaluationbenchmark} dataset is presented in Tab.~\ref{tab:MME}. Our method achieves better overall (Total) results and equal or improved performance in existence and counting, demonstrating its effectiveness in object recognition. However, it does not improve position accuracy and exhibits varying behavior on color. This diversity may stem from the inherent capabilities of MLLMs, which cannot be solely enhanced through token influence balancing.

\label{sec:MME}
\begin{table}[htbp]
\centering
\vspace{-0.3cm}
\caption{Result on MME Dataset.}
\vspace{-0.2cm}
\footnotesize
\begin{tabularx}{\textwidth}{l*{7}{X}}
\toprule
MLLMs & Method & Existence $\uparrow$ & Count $\uparrow$ & Position $\uparrow$ & Color $\uparrow$ & Total $\uparrow$\\
\midrule
\multirow{2}{*}{LLaVA-v1.5} & base & 190.0 & 140.0 & 128.3 & 155.0 & 613.3 \\
& ours & 190.0 & \textbf{153.3} & 128.3 & \textbf{163.3} & \textbf{634.9} \\
\midrule
\multirow{2}{*}{IntructBLIP} & base & 180.0 & 55.0 & 50.0  & 130.0 & 415.0 \\
& ours & \textbf{185.0} & 55.0 & 50.0 & 130.0 & \textbf{420.0} \\
\midrule
\multirow{2}{*}{mPLUG-Owl2} & base & 170.0 & 145.0 & 73.3 & \textbf{158.3} & 546.6 \\
& ours & 170.0 & \textbf{150.0} & 73.3 & 150.0 & \textbf{548.3} \\
\bottomrule
\end{tabularx}
\label{tab:MME}
\vspace{-0.4cm}
\end{table}

\section{Other Results of POPE}
\label{sec:POPE_2}

\begin{table}[htbp]
\centering
\caption{More Results on POPE \cite{li2023evaluating}.}
\vspace{-0.2cm}
\footnotesize
\begin{tabularx}{\textwidth}{l*{9}{X}}
\toprule
\multirow{2}{*}{Dataset} & \multirow{2}{*}{Setting} & \multirow{2}{*}{Method} & \multicolumn{2}{c}{LLaVA-v1.5} & \multicolumn{2}{c}{InstructBLIP} & \multicolumn{2}{c}{mPLUG-Owl2} \\
\cmidrule(lr){4-5} \cmidrule(lr){6-7} \cmidrule(lr){8-9} 
& & & Acc $\uparrow$ & F1 $\uparrow$ & Acc $\uparrow$ & F1 $\uparrow$ & Acc $\uparrow$ & F1 $\uparrow$ \\
\midrule
\multirow{4}{*}{MSCOCO} & \multirow{2}{*}{Random} & base  & 87.1 & 85.4 & 87.1 & 85.7 & 86.0 & 84.4 \\ 
 & & ours & \textbf{87.4} & \textbf{86.0} & \textbf{87.9} & \textbf{86.8} & \textbf{87.9} & \textbf{87.1} \\ 
 \cmidrule(lr){2-9} 
 & \multirow{2}{*}{Popular} & base  & 85.9 & 84.4 & 84.2 & 83.6 & 84.6 & 83.2 \\ 
 &  & ours & \textbf{86.2} & \textbf{84.8} & \textbf{85.0} & \textbf{84.3} & \textbf{86.4} & \textbf{85.7} \\ 
 \midrule
 \multirow{6}{*}{A-OKVQA} & \multirow{2}{*}{Random} & base  & 88.0 & \textbf{87.6} & 88.5 & 88.5 & 86.5 & 85.7 \\ 
 & & ours & \textbf{88.1} & 87.4 & \textbf{88.8} & \textbf{88.8} & \textbf{88.4} & \textbf{88.1} \\ 
 \cmidrule(lr){2-9} 
 & \multirow{2}{*}{Popular} & base  & 85.5 & 85.1 & 81.9 & 83.1 & 82.4 & 82.2 \\ 
 &  & ours & 85.5 & 85.1 & \textbf{82.3} & \textbf{83.4} & \textbf{85.1} & \textbf{85.3} \\ 
\cmidrule(lr){2-9} 
 & \multirow{2}{*}{Adversarial} & base  & 79.1 & 79.9& 74.8 & 77.9 & 74.7 & 76.9 \\
 &  & ours & \textbf{79.5} & \textbf{80.1} & \textbf{75.3} & \textbf{78.2} & \textbf{78.2} & \textbf{79.9} \\ 
\midrule
 \multirow{6}{*}{GQA} & \multirow{2}{*}{Random} & base  & 88.9 & 88.2&87.2 & 87.1 & 85.2 & 84.0 \\ 
 & & ours & 88.9 & 88.2 & 87.2 & \textbf{87.2} & \textbf{86.1} & \textbf{85.0} \\ 
 \cmidrule(lr){2-9} 
 & \multirow{2}{*}{Popular} & base   & 84.1 & 84.1 & 78.6 & 80.4 & 78.7 & 78.5 \\ 
 &  & ours & \textbf{84.2} & 84.1 & \textbf{78.8} & 80.4 & \textbf{81.0} & \textbf{80.5} \\ 
\cmidrule(lr){2-9} 
 & \multirow{2}{*}{Adversarial} & base   & 80.8 & 81.3 & 75.9 & 78.4 & 76.4 & 76.8 \\ 
 &  & ours & \textbf{81.1} & \textbf{81.6} & \textbf{76.1} &\textbf{ 78.5} & \textbf{79.2} & \textbf{79.1} \\ 
\bottomrule
\end{tabularx}
\label{tab:POPE_2}
\end{table}

We report our experimental results on the POPE dataset, in addition to MSCOCO and adversarial settings, in Tab.~\ref{tab:POPE_2}. The results indicate that our method improves performance across all baseline MLLMs, with more significant gains observed in the adversarial setting.
This discrepancy likely arises because adversarial scenarios require models to rely more heavily on visual inputs, aligning with our method's focus on enhancing visual influence. Conversely, for popular and random objects, textual data often provides sufficient statistical information, reducing the necessity for increased visual input reliance.

\section{Question Category Results on the AMBER Dataset}

We report discriminative results across different question categories in Tab.~\ref{tab:AMBER_d}. Our method improves performance in nearly all categories across all MLLMs. The improvement in InternVL2’s object existence is minor, likely due to its already high visual influence ratio. For LLaVA-v1.5 and mPLUG-Owl2, which have lower original visual influence ratios, our method achieves more substantial gains in existence, attribute, and state categories.

\begin{table}[htbp]
\centering
\caption{Results on the Question Categories of Discriminative Task on AMBER Dataset.}
\vspace{-0.2cm}
\footnotesize
\begin{tabularx}{\textwidth}{l l *{10}{X}}
\toprule
\multirow{2}{*}{Category} & \multirow{2}{*}{Metric} & \multicolumn{2}{c}{InstructBLIP} & \multicolumn{2}{c}{LLaVA-v1.5} & \multicolumn{2}{c}{LLaVA-v1.6} & \multicolumn{2}{c}{mPLUG-Owl2} & \multicolumn{2}{c}{Intern-VL2} \\
&&base&ours&base&ours&base&ours&base&ours&base&ours \\
\midrule
\multirow{4}{*}{Existence} & acc & 70.0 & 79.8 & 70.8 & 93.2 & 92.9 &93.0 &75.2 &89.9 & 90.6 &90.6 \\
& P & 100.0 & 100.0 & 100.0 & 100.0 & 100.0 & 100.0 & 100.0 & 100.0 & 100.0 & 100.0 \\
& R & 70.0 &79.8 & 70.8 & 93.2 & 92.9 &93.0 & 75.2 & 89.9 & 90.6 & 90.6 \\
& F1 & 82.3 & 88.7 &82.9 & 96.4 & 96.3 & 96.3 & 85.8 & 94.6 & 95.0 & 95.0 \\
\midrule
\multirow{4}{*}{Attribute} & acc &71.9 & 78.3 & 72.3 & 76.1 &75.2 & 77.1 & 73.9 & 78.2 &82.3 & 82.6\\
& P & 76.0 & 81.7 & 87.3 & 74.0 & 74.6 & 76.4 &86.0 &76.9 & 80.9 & 80.9\\
& R & 64.3 & 73.0 & 52.2 & 82.7 & 83.0 & 83.9 & 57.1 & 81.8 & 84.7 &85.2 \\
& F1 & 69.7 & 77.1 & 65.3 &78.1 & 78.5 & 80.0 & 68.6 & 79.3 & 82.8 &83.0 \\
\midrule
\multirow{4}{*}{State} & acc & 73.4 & 76.4 & 68.2 & 73.3 & 78.6 & 75.2 & 70.5 & 77.9 & 81.2 &81.2 \\
& P & 75.1 & 77.1 &86.2 & 70.3 & 78.6 & 74.7 & 84.9 & 75.5 & 79.1 & 78.7\\
& R & 70.6 & 75.3 & 43.3 & 82.0 & 78.5 & 82.9 & 49.8 &83.1 &84.8 &85.5\\
& F1 & 72.8 & 76.2 & 57.6 & 75.7 & 78.5 & 78.6 & 62.8 & 79.1 & 81.8 & 82.0\\
\midrule
\multirow{4}{*}{Number} & acc & 65.4 & 80.6 &75.0 & 80.1 & 80.1 & 80.2 & 77.8 & 76.5 & 82.6 & 83.3\\
& P & 75.4 & 93.1 & 86.9 & 79.1 & 79.2 & 78.6 & 86.0 & 77.0 & 83.0 & 84.0\\
& R & 45.8 & 66.2 & 59.5 & 82.4 & 81.7 & 84.4 & 66.9 & 77.0 & 82.0 & 82.3\\
& F1 & 57.0 & 77.4 & 70.6 & 80.7 & 80.4 & 81.4 & 75.3 & 77.0 & 82.5 & 83.1\\
\midrule
\multirow{4}{*}{Action} & acc & 79.7 &83.7 & 83.6 & 82.3 & 81.9 &80.4 &84.0 & 84.1 &88.4 &88.6\\
& P & 82.5 &88.5 & 92.9 &85.9 &79.4 &81.2 & 90.9 & 85.9 & 86.5 & 86.8\\
& R & 75.3 & 77.5 & 72.7 &87.4 & 86.0 & 88.6 & 75.5 & 85.9 & 90.9 &91.2\\
& F1 & 78.7 & 82.6 & 81.6 & 86.6 & 82.6 & 84.7 & 82.5 & 85.9 & 88.6 & 88.9\\
\midrule
\multirow{4}{*}{Relation} & acc & 62.7 & 71.9 & 71.8 &61.5 &64.5 &65.7 & 70.5 & 76.9 & 72.1 & 77.0\\
& P & 56.2 & 64.0 & 65.9 & 51.9 & 54.0 & 56.6 & 61.0 & 67.9 & 60.0 & 65.1\\
& R & 48.6 & 73.4 & 66.3 & 97.7 & 95.1 & 87.2 & 79.5 & 83.9 & 98.3 & 95.6\\
& F1 & 52.1 & 68.4 & 66.1 & 67.8 & 68.9 & 68.6 & 69.0 & 75.1 & 74.5 & 77.5\\
\bottomrule
\end{tabularx}
\label{tab:AMBER_d}
\vspace{-0.4cm}
\end{table}

\section{Different Sampling Strategies}
\label{sec:Sampling}
Tab.~\ref{tab:sampling} presents an ablation study on sampling strategies (non-greedy vs. greedy). 
We follow the non-greedy sampling setting of VCD \cite{leng2024mitigating}, where both top-p and temperature are set to 1. As shown, our method consistently improves performance across both sampling strategies.


\begin{table}[htbp]
\centering
\caption{Ablation Study on Sampling Strategies on POPE MSCOCO Adversarial Dataset.}
\footnotesize
\begin{tabularx}{\textwidth}{*{10}{X}}
\toprule
\multirow{2}{*}{strategy} 
&\multirow{2}{*}{Method} & \multicolumn{2}{c}{LLaVA-v1.5} & \multicolumn{2}{c}{InstructBLIP} & \multicolumn{2}{c}{mPLUG-Owl2} \\
\cmidrule(lr){3-4} \cmidrule(lr){5-6} \cmidrule(lr){7-8} 
&& Acc & F1  & Acc & F1 & Acc & F1 \\
\midrule
\multirow{2}{*}{non-greedy} &base& $79.0_{\pm 0.51}$ & $81.1_{\pm 0.53}$ & $71.6_{\pm 0.49}$ & $74.7_{\pm 0.46}$ & $71.5_{\pm 0.30}$  & $76.6_{\pm 0.28}$ \\
&ours& \textbf{82.3}$_{\pm 0.27}$ & 81.1$_{\pm 0.31}$ & \textbf{82.2}$_{\pm 0.29}$ & \textbf{81.8}$_{\pm 0.25}$ & \textbf{83.2}$_{\pm 0.27}$  & \textbf{82.9}$_{\pm 0.26}$ \\
\midrule
\multirow{2}{*}{greedy}&base &80.9 & 81.6 & 79.8 & 81.4 & 72.5 & 77.5 \\
&ours & \textbf{83.5} & \textbf{82.1} & \textbf{82.5} & \textbf{82.1} & \textbf{84.2}  & \textbf{83.7} \\
\bottomrule
\end{tabularx}
\label{tab:sampling}
\end{table}

\section{Different Model Size}
\label{sec:size}
We evaluate our method on different model sizes, 7B and 13B, for LLaVA-v1.5 and InstructBLIP, as shown in Tab.~\ref{tab:size}. The results indicate consistent improvements across various model sizes. In each model series, the smaller model gets a larger performance boost. With our method, we can achieve high accuracy and detection rates with a smaller 7B model, outperforming a 13B model at its original performance level.

\begin{table}[htbp]
\centering
\caption{Ablation Study on Model Size on LLaVA-QA90 Dataset.}
\footnotesize
\begin{tabularx}{\textwidth}{*{9}{X}}
\toprule
\multirow{3}{*}{Mothod} & \multicolumn{4}{c}{LLaVA-v1.5} & \multicolumn{4}{c}{InstructBLIP} \\
\cmidrule(lr){2-5} \cmidrule(lr){6-9} 
& \multicolumn{2}{c}{7B} & \multicolumn{2}{c}{13B} &\multicolumn{2}{c}{7B} & \multicolumn{2}{c}{13B} \\
\cmidrule(lr){2-3} \cmidrule(lr){4-5} \cmidrule(lr){6-7} \cmidrule(lr){8-9}
&  Acc & Det & Acc & Det  & Acc & Det  & Acc & Det  \\
\midrule
base& 3.23 & 3.54 & 4.78 & 4.2 & 3.84  & 4.07  & 5.67 & 4.88  \\
ours & \textbf{6.20} & \textbf{5.14} & \textbf{7.36} & \textbf{6.5} & \textbf{6.28}  & \textbf{4.77} & \textbf{6.42} & \textbf{5.99} \\
\bottomrule
\end{tabularx}
\label{tab:size}
\end{table}

\section{Revisiting the Accuracy–Informativeness Trade-off}
In the main paper, we report recall and output length alongside CHAIR scores, since our objective is to evaluate models under a balance of \emph{accuracy} and \emph{informativeness}. This choice is deliberate: our early stopping mechanism can be tuned to shorten responses, which naturally reduces CHAIR scores but at the expense of recall and content richness. Consequently, the trade-off introduced by early stopping is an explicit design choice, and it can be adjusted depending on the requirements of a specific application.  

Direct comparison with SOTA methods that omit recall and generation length is therefore not entirely fair. Our analysis confirms that CHAIR scores can drop substantially when outputs are truncated, underscoring the importance of jointly reporting recall and length to present a complete view of performance. Without these complementary metrics, lower CHAIR values may simply reflect shorter, less informative responses rather than genuine improvements in visual grounding.  
To enable a fairer comparison with prior work, we adjust our early stopping threshold to $12\%$. Under this setting, our method achieves lower CHAIR scores while maintaining competitive recall, thereby outperforming both approaches. This demonstrates that our framework not only mitigates hallucination effectively but also preserves informativeness. Moreover, the adjustable nature of the early stopping mechanism ensures that users can flexibly select the optimal balance between accuracy and informativeness for their specific use cases.

\begin{table}[htbp]
\centering
\caption{Comparison with SOTA Methods with $12\%$ Early Stopping Threshold.}
\footnotesize
\begin{tabular}{l l c c c c}
\toprule
MLLM & Method & $C_s$ $\downarrow$ & $C_i$ $\downarrow$ &R $\uparrow$ & Len $\uparrow$ \\
\hline
\multirow{5}{*}{LLaVA-v1.5} & base        & 48.8 & 13.4 & \textbf{78.6} & \textbf{99.8} \\
& PAI~\cite{liu2024paying}       & 24.8 & 6.9  & -    & -    \\
& Middle~\cite{jiang2025devils}      & 25.0 & 6.7  & -    & -    \\
& Ours\_ES\_12\% & \textbf{23.5} & \textbf{6.5} & 55.0 & 54.1 \\
\hline
\end{tabular}
\label{tab:trade0ff_study}
\end{table}

\section{Grouping Multiple Influential Tokens with Respect to the Anchor Object}
We also explored top-$2$ variants as a straightforward extension, but they did not yield consistent improvements over the top-$1$ design. One possible reason is that, for some predicted tokens, supervision is effectively dominated by a single visual token, so forcing a top-$2$ aggregation can dilute this primary contribution rather than help. As a result, the top-$2$ scheme may only benefit a subset of objects that are genuinely supported by multiple visual tokens, leading to limited overall gains.

\begin{table}[htbp]
\centering
\caption{Top-k Anchor-Object Influential Token Strategies on the POPE MSCOCO Adversarial Dataset}
\footnotesize
\begin{tabularx}{\textwidth}{*{9}{X}}
\toprule
\multirow{2}{*}{Norm} & \multicolumn{2}{c}{LLaVA-v1.5} & \multicolumn{2}{c}{InstructBLIP} & \multicolumn{2}{c}{mPLUG-Owl2} \\
\cmidrule(lr){2-3} \cmidrule(lr){4-5} \cmidrule(lr){6-7} 
& Acc & F1  & Acc & F1 & Acc & F1 \\
\midrule
top-$1$ & 83.5 & 82.1 & 82.5 & 82.1 & 84.2  & 83.7 \\
top-$2$ &  83.1 &  82.8 &  82.9 & 82.5  &  82.3  & 82.6  \\
\bottomrule
\end{tabularx}
\label{tab:norm}
\vspace{-0.2cm}
\end{table}

\section{Hyper parameter Study}
\label{sec:hyper}
\noindent\textbf{Maximum $\alpha_m$} serves primarily as a \textit{precautionary} upper bound. Because Eq.~13 in the main text naturally bounds $\alpha$, this maximum threshold is rarely triggered, as evidenced by the low clipping rates reported in Tab.~\ref{tab:study_alpha_r}. 

\begin{table}[htbp]
\centering
\caption{Clipping Rate of $\alpha_m$ at Maximum Value}
\footnotesize
\begin{tabularx}{\textwidth}{p{3cm}*{2}{X}}
\toprule
Task & Discriminative @ POPE Adversarial Setting & Generation @ MSCOCO Subset \\
\midrule
GACD (LLaVA-v1.5) & 0.04\% & 0.07\% \\
\bottomrule
\end{tabularx}
\label{tab:study_alpha_r}
\end{table}

Nevertheless, to determine the optimal value for $\alpha_m$ and assess its impact on model performance, we conducted a hyperparameter search using LLaVA-v1.5. For discriminative tasks, we evaluated $\alpha_m \in \{1, \dots, 6\}$ on the POPE dataset. For open-ended generation tasks on a subset of MSCOCO \cite{deng2024seeing}, we observed garbled text outputs when $\alpha_m \ge 5$; thus, we restricted our search space to $\alpha_m \in \{1, \dots, 4\}$. 
As shown in Tab.~\ref{tab:study_alpha_d} and Tab.~\ref{tab:study_alpha_g}, performance on the POPE discriminative task is relatively insensitive to variations in $\alpha_m$. Conversely, performance on the generative task initially improves as $\alpha_m$ increases but degrades at higher values. Based on these findings, we set the optimal maximum amplification factor $\alpha_m$ to 5 for discriminative tasks and 3 for generative tasks across all our experiments.

\begin{table}[htbp]
\centering
\caption{$\alpha_m$ Study For Discriminative Task On POPE \cite{li2023evaluating} in MSCOCO Adversarial Setting.}
\footnotesize
\begin{tabularx}{\textwidth}{p{2cm} *{13}{X}}
\toprule
\multirow{2}{*}{Maximum $\alpha_m$} & \multicolumn{2}{c}{1} & \multicolumn{2}{c}{2} & \multicolumn{2}{c}{3} &\multicolumn{2}{c}{4} &\multicolumn{2}{c}{5} &\multicolumn{2}{c}{6}\\
\cmidrule(lr){2-3} \cmidrule(lr){4-5} \cmidrule(lr){6-7} \cmidrule(lr){8-9}  \cmidrule(lr){10-11} \cmidrule(lr){12-13} 
& Acc $\uparrow$ & F1 $\uparrow$ & Acc $\uparrow$ & F1 $\uparrow$ & Acc $\uparrow$ & F1 $\uparrow$ & Acc $\uparrow$ & F1 $\uparrow$ & Acc $\uparrow$ & F1 $\uparrow$ & Acc $\uparrow$ & F1 $\uparrow$ \\
\midrule
LLaVA-v1.5 & 83.4 & 82.0 & 83.4 & 82.0 & 83.4 & 82.0 & 83.4 & 82.0& 83.5 & 82.1 &83.4 & 82.1\\
\bottomrule
\end{tabularx}
\label{tab:study_alpha_d}
\end{table}

\begin{table}[htbp]
\centering
\vspace{-0.2cm}
\caption{Maximum $\alpha_m$ Study For Generation Task on the MSCOCO Subset.}
\footnotesize
\begin{tabularx}{\textwidth}{l*{18}{X}}
\toprule
\multirow{2}{*}{Maximum $\alpha_m$} & \multicolumn{4}{c}{1} & \multicolumn{4}{c}{2} & \multicolumn{4}{c}{3} &\multicolumn{4}{c}{4}\\
\cmidrule(lr){2-5} \cmidrule(lr){6-9} \cmidrule(lr){10-13} \cmidrule(lr){14-17} 
& $C_S \downarrow$& $C_I \downarrow$ & $R \uparrow$ & $Len$ & $C_S \downarrow$& $C_I \downarrow$ & $R \uparrow$ & $Len$ & $C_S \downarrow$& $C_I \downarrow$ & $R \uparrow$ & $Len$ & $C_S \downarrow$& $C_I \downarrow$ & $R \uparrow$ & $Len$\\
\midrule
LLaVA-v1.5 & 44.0 & 11.8& 76.2 & 86.1 & 41.4& 11.1& 77.4 & 84.8 & 41.0 & 10.9 & 77.3 & 85.0 & 41.4 &  10.9& 77.3 & 84.9 \\
\bottomrule
\end{tabularx}
\label{tab:study_alpha_g}
\end{table}

\noindent\textbf{Early Stopping Threshold}. 
Our study follows a systematic, data-driven protocol. 
We conducted a grid search on a subset of the MSCOCO dataset subset following \cite{deng2024seeing}. Recognizing that the visual influence ratio varies across models, we first compute the mean and variance of the EOS visual ratio and the subsequent hallucination rate on a small calibration set. This statistical window defines the range and step size for this study.
By searching over the corresponding range, we show results in Tab.~\ref{tab:study_es}. These results demonstrate that varying the ES threshold primarily mediates the trade-off between recall and hallucination rate.
Our goal is to have balanced recall ($R$) and instance-level hallucination ($C_I$), leading us to select thresholds of $7\%$ for LLaVA-v1.5 and LLaVA-v1.6, $25\%$ for InstructBLIP, $2.5\%$ for mPLUG-Owl2 and $10\%$ for InternVL2.
We additionally ran an experiment to measure the ES activation rate using LLaVA-v1.5 with ES threshold $7\%$. As shown in Tab.~\ref{tab:study_es_r}, ES fires on only $8.7\%$ of the test samples, and when it does, the generated responses are on average just 0.7 tokens shorter. This indicates that ES rarely, and only minimally, truncates outputs.

\begin{table}[htbp]
\centering
\caption{Early Stopping Threshold Study on the MSCOCO Subset}
\footnotesize
\begin{tabularx}{\textwidth}{p{0.2cm}*{21}{X}} 
\toprule
\multirow{2}{*}{} & \multicolumn{4}{c}{LLaVA-v1.5} & \multicolumn{4}{c}{LLaVA-v1.6} & \multicolumn{4}{c}{IntructBLIP} &\multicolumn{4}{c}{mPLUG-Owl2} &\multicolumn{4}{c}{InternVL2}\\
\cmidrule(lr){2-5} \cmidrule(lr){6-9} \cmidrule(lr){10-13} \cmidrule(lr){14-17} \cmidrule(lr){18-21} 
& $6\%$ & $7\%$ & $8\%$ & $9\%$ &$5\%$ & $7\%$ & $9\%$ & $11\%$ & $15\%$ & $20\%$ & $25\%$ & $30\%$ & $2\%$ & $2.5\%$ & $3\%$ & $3.5\%$ & $8\%$ & $10\%$ & $12\%$ & $14\%$\\
\midrule
$C_S$ &45.6&41.0&36.6&31.4&29.0&26.0&23.0&17.8&52.6&51.4&47.4&36.0&51.8&45.0&41.2&41.0&37.0&35.2&34.6&32.9 \\
$C_I$ &11.5&10.9&10.2&10.2&8.5&8.1&7.8&7.5&15.0&14.3&13.4&11.7&13.7&12.4&11.0&11.0&8.6&8.1&8.0&7.9 \\
$R$ &79.7&77.3&75.2&70.8&68.5&63.0&58.8&53.0&75.1&74.4&72.3&68.8&77.7&74.9&73.8&73.5&65.8&65.4&65.4&64.0 \\
$L_{en}$ &92.$_0$&85.$_0$&75.$_4$&63.$_9$&119.$_1$&101.$_8$&81.$_1$&62.$_7$&107.$_9$&103.$_4$&93.$_9$&74.$_3$&89.$_1$&83.$_5$&78.$_9$&77.$_8$&180.$_4$&175.$_5$&170.$_6$&162.$_2$ \\
\bottomrule
\end{tabularx}
\label{tab:study_es}
\vspace{-0.5cm}
\end{table}

\begin{table}[htbp]
\centering
\caption{Activation Rate of the Early Stopping on the MSCOCO Subset}
\footnotesize
\begin{tabularx}{\textwidth}{p{3cm}*{3}{X}}
\toprule
\multicolumn{2}{c}{Methods} & Activate Percentage & Average Length \\
\midrule
\multirow{2}{*}{LLaVA-v1.5 ($7\%$)} & base & - & 85.1 \\
& ours & $8.7\%$ & 84.4 \\
\bottomrule
\end{tabularx}
\label{tab:study_es_r}
\vspace{-0.3cm}
\end{table}

\section{Confidence and Visual Influence}

\begin{figure*}[!htbp]
    \centering
    \begin{subfigure}[b]{0.48\textwidth}
        \includegraphics[width=\textwidth]{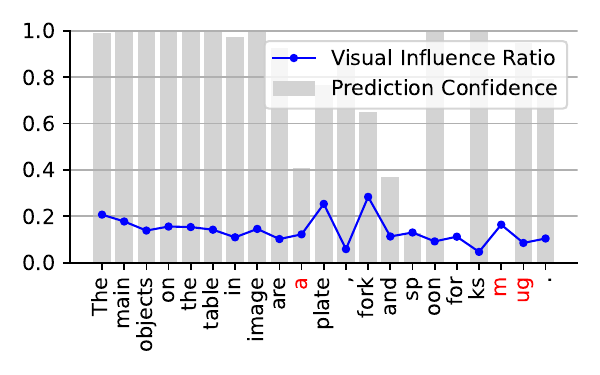}
        \vspace{-0.8cm}
        \caption{mPLUGOwl2}
        \label{fig:conf_ori}
    \end{subfigure}
    \hfill
    \begin{subfigure}[b]{0.48\textwidth}
        \includegraphics[width=\textwidth]{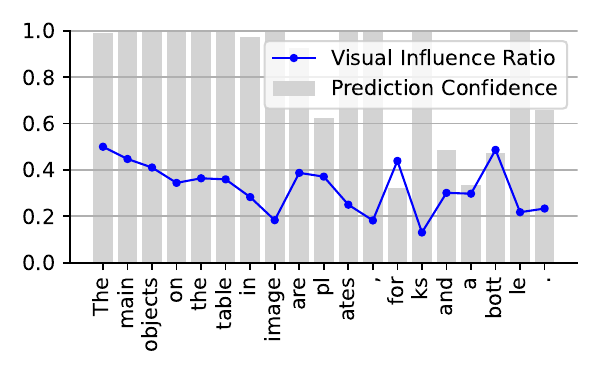}
        \vspace{-0.8cm}
        \caption{mPLUGOwl2-GACD}
        \label{fig:conf_our}
    \end{subfigure}
    \vspace{-0.2cm}
    \caption{Comparison of prediction confidence with and without GACD. (a) Without GACD, mPLUGOwl2 exhibits low confidence in \textcolor{red}{hallucinated predictions} and near-zero confidence in the initial predictions for `forks' and `mug'. (b) With GACD, mPLUGOwl2's confidence increases alongside the \textcolor{blue}{visual influence ratio}, effectively mitigating hallucinations.}
    \label{fig:confidence}
    \vspace{-0.4cm}
\end{figure*}

Low confidence often signals potential failure modes in base MLLMs. Here, we demonstrate that our method not only improves accuracy but also increases model confidence. It remains effective even in low-confidence regions for three reasons: 1) We aggregate token gradients at the component level (Eq.~\ref{eq:group-influence-emitted}) rather than using individual token gradients which yields robustness against local gradient noise. 2) We adjust influence towards visual tokens which consistently reduces the hallucination likelihood; 3) Empirically, low model confidence does not correlate with noisy gradients. In our experiments, pretrained MLLMs usually maintain meaningful gradient signals even at low confidence levels.
Fig.~\ref{fig:confidence} shows an example where the baseline model mPLUG-Owl2 exhibits low confidence in hallucinated predictions and near-zero confidence in the initial predictions for `forks' and `mug'. With GACD, prediction confidence increases alongside the visual influence ratio, with the minimum confidence rising to over $30\%$.

\section{Additional Implementation and Experimental Details}
\label{sec:ED}
We identify noun tokens using the spaCy library
due to its lightweight, CPU-only operation; this component is interchangeable with any alternative noun detector. At each step, if the current token is classified as a noun and at least one noun has appeared in previous outputs, we trigger noun-only grouping. In this mode, we inject noun-related visual tokens into the negative guidance; otherwise, negative guidance is computed using only text tokens.
For experiments on the Amber dataset \cite{wang2023llm}, we adopt the original data splits and evaluation metrics. In the MSCOCO \cite{lin2014microsoft} subset, we follow the data partitioning and evaluation protocol of Deng et al.~\cite{deng2024seeing}, with splits available in their official repository. For the LLaVA-QA90 \cite{liu2024visual}, MME \cite{fu2024mmecomprehensiveevaluationbenchmark} and POPE \cite{li2023evaluating} datasets, our setup replicates that of Leng et al.~\cite{leng2024mitigating} and use their provided scoring scripts for LLaVA-QA90. 
Experiments on MMBench \cite{liu2025mmbench}, MM-Vet \cite{yu2023mm} follow the \href{https://github.com/OpenGVLab/VLMEvalKit_InternVL2_5}{VLMEvalKit\_InternVL2\_5} repository.
All comparison methods are executed using their official code; we only modify them to enforce greedy sampling and a uniform maximum generation length to align with our experimental settings.

\section{Influence Ratio in VQA}
\label{sec:VQA}
\begin{figure}[H]
    \centering
    \begin{subfigure}[b]{\textwidth}
        \hspace*{5.5cm}
        \includegraphics[width=0.2\textwidth]{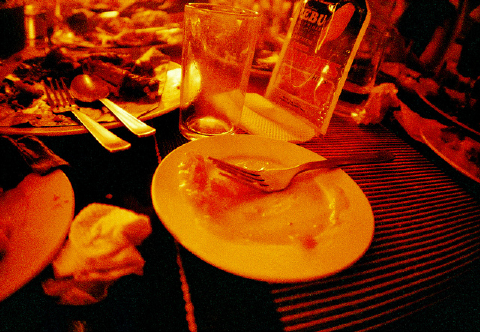}
        \caption{What are the main objects on the table in the image?}
        \label{fig:input_qa}
    \end{subfigure}
    \begin{minipage}[b]{0.85\textwidth}
        \begin{subfigure}[b]{0.5\textwidth}
            \includegraphics[width=\textwidth]{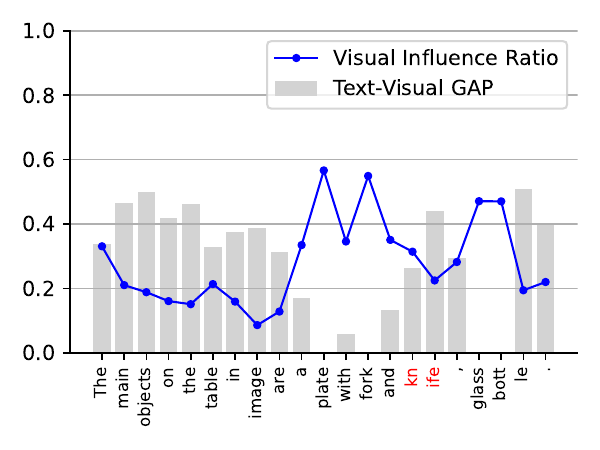}
            \caption{LLaVAv1.5}
            \label{fig:vqa_llava_ori}
        \end{subfigure}
        \hfill
        \begin{subfigure}[b]{0.5\textwidth}
            \includegraphics[width=\textwidth]{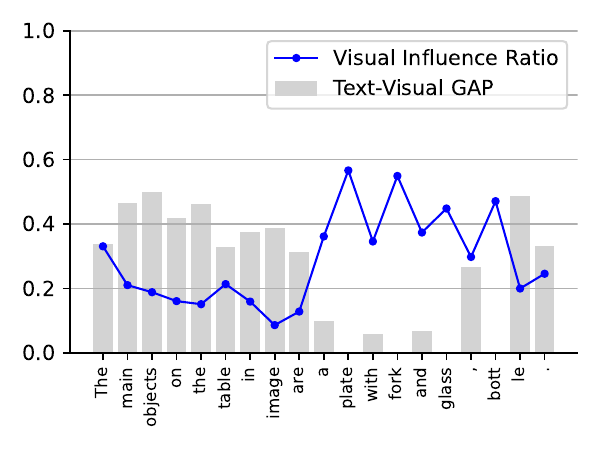}
            \caption{LLaVAv1.5-GACD}
            \label{fig:vqa_llava_ours}
        \end{subfigure}
    \end{minipage}
        \begin{minipage}[b]{0.85\textwidth}
        \begin{subfigure}[b]{0.5\textwidth}
            \includegraphics[width=\textwidth]{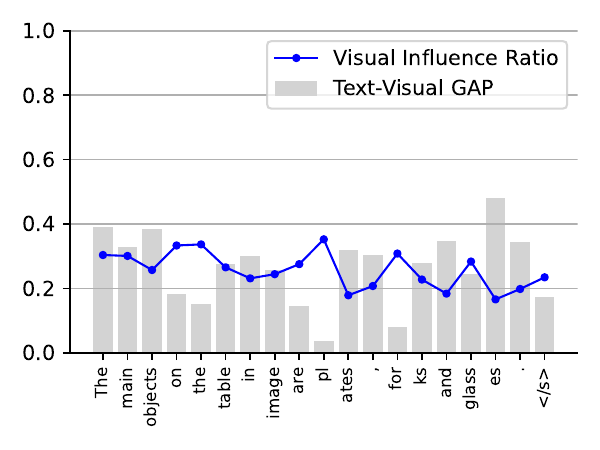}
            \caption{InstructBLIP}
            \label{fig:vqa_blip_ori}
        \end{subfigure}
        \hfill
        \begin{subfigure}[b]{0.5\textwidth}
            \includegraphics[width=\textwidth]{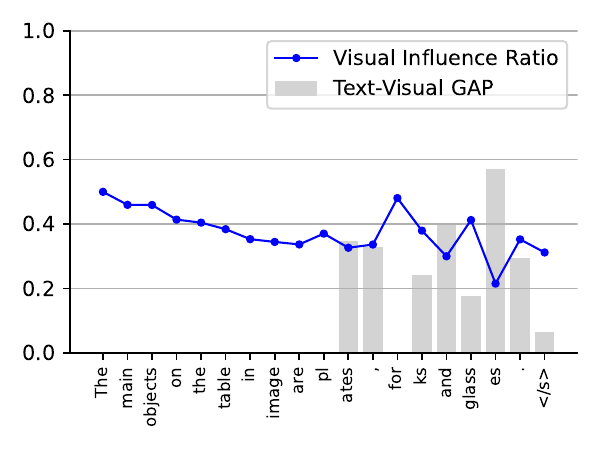}
            \caption{InstructBLIP-GACD}
            \label{fig:vqa_blip_ours}
        \end{subfigure}
    \end{minipage}
        \begin{minipage}[b]{0.85\textwidth}
        \begin{subfigure}[b]{0.5\textwidth}
            \includegraphics[width=\textwidth]{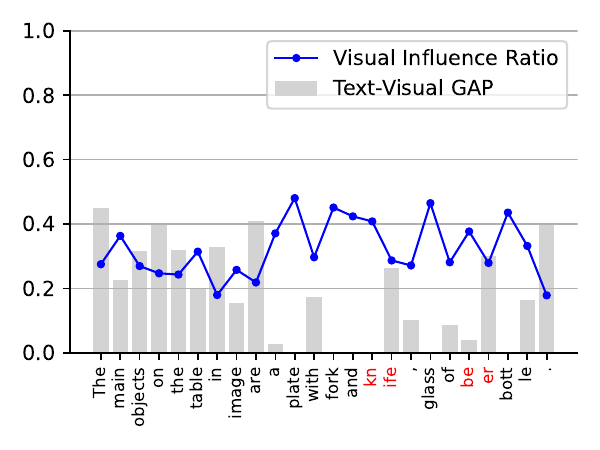}
            \caption{InternVL2}
            \label{fig:vqa_internvl_ori}
        \end{subfigure}
        \hfill
        \begin{subfigure}[b]{0.5\textwidth}
            \includegraphics[width=\textwidth]{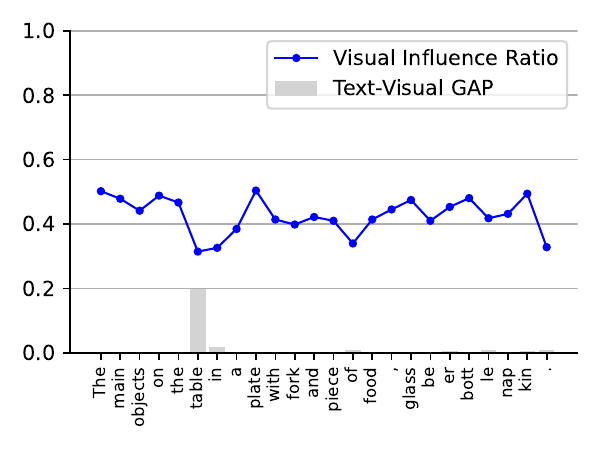}
            \caption{InternVL2-GACD}
            \label{fig:vqa_internvl_ours}
        \end{subfigure}
    \end{minipage}
    \caption{Influence Ratio across Predicted Tokens in VQA: (left) Baseline predictions; (right) Predictions with GACD. 
    GACD effectively mitigate Text-Visual GAP, balancing text-visual bias.
    (f) The original InternVL2 shows a dominant visual influence ratio at the hallucinated prediction \textcolor{red}{`knife'}, indicating a co-occurrence bias that remains unaddressed even with dominant visual influence. (g) GACD successfully eliminates co-occurrence hallucinations, including \textcolor{red}{`knife'}.
    }
    \label{fig:ratio}
\end{figure}

Fig.~\ref{fig:ratio} illustrates the visual influence ratio across outputs in VQA tasks, comparing baseline predictions with those obtained after applying GACD. 
The results confirm that text tokens dominate influence across MLLMs, including InternVL2, which exhibits a relatively high visual influence ratio. As shown in Fig.~3 of the main paper, the overall $60\%$–$100\%$ visual influence ratio across the POPE dataset suggests that visual inputs predominantly determine object existence in VQA tasks.
GACD enhances visual influence, effectively balancing text-visual bias. 
Furthermore, the visualization on InternVL2 demonstrates that the co-occurrence hallucination `knife' persists despite a high visual influence. GACD successfully eliminates this co-occurrence hallucination.

\section{Influence Ratio in Image Caption}
\label{sec:IC}
\begin{figure*}[!htbp]
    \centering
    \includegraphics[width=\textwidth]{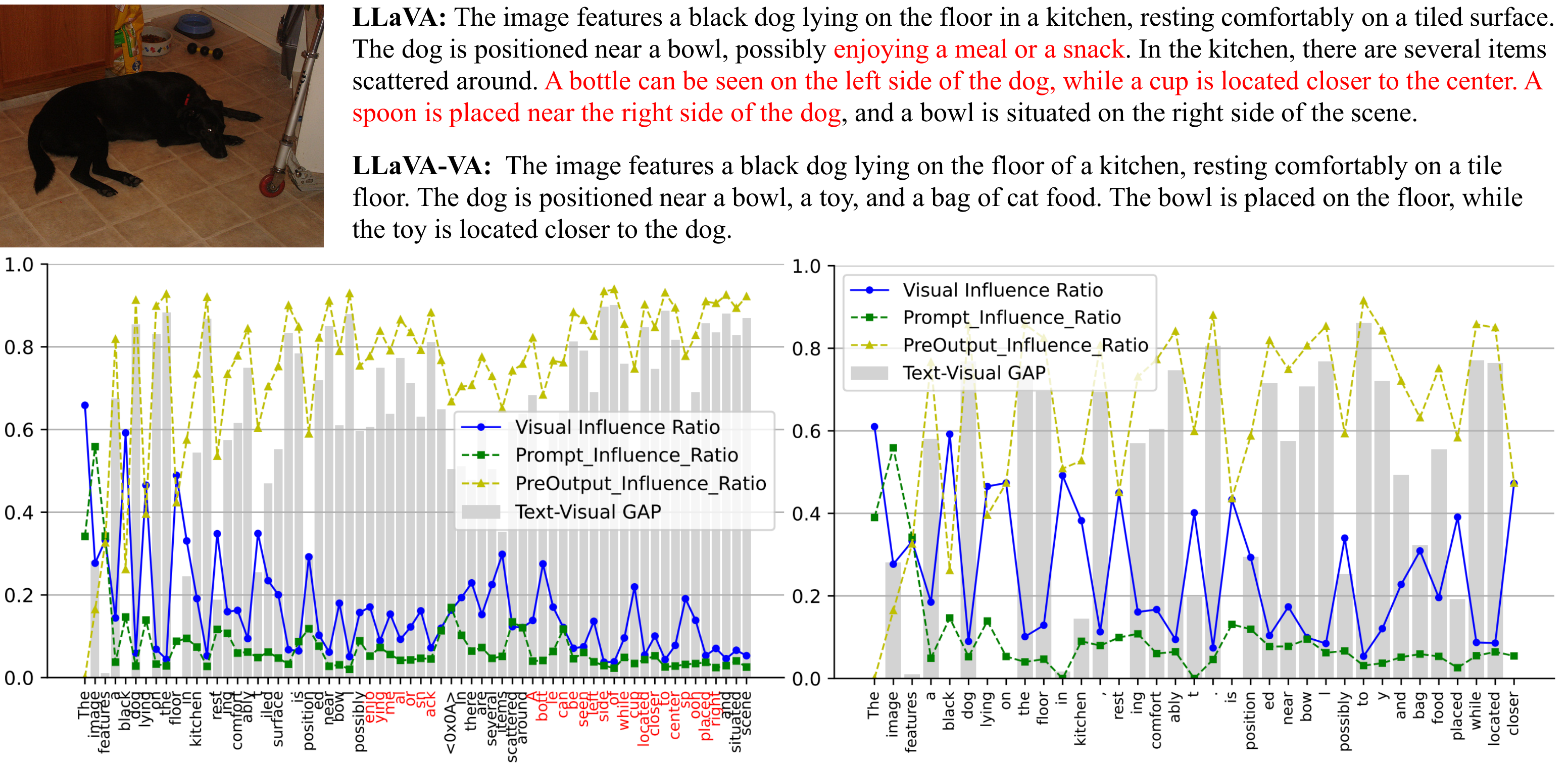}
    \vspace{-0.6cm}
    \caption{Comparison of influence ratios across predicted tokens with and without GACD. (Left) Without GACD, the influence gap between \textcolor{myyellow}{previous output tokens} and \textcolor{blue}{visual tokens} widens as more tokens are generated. (Right) With GACD, the gap is periodically narrowed to nearly zero, mitigating this trend and reducing hallucination.}
    \vspace{-0.4cm}
    \label{fig:IC}
\end{figure*}

We further visualize the influence ratio in the image captioning task. 
Fig.~\ref{fig:IC} shows that in the baseline LLaVA-v1.5, the influence gap between previous output tokens and visual tokens widens as more tokens are generated. However, GACD effectively narrows this gap, preventing visual information from being forgotten and thereby reducing hallucinations.

\section{Qualitative Example on Occluded Images}
\begin{figure*}[!htbp]
    \centering
    \includegraphics[width=\textwidth]{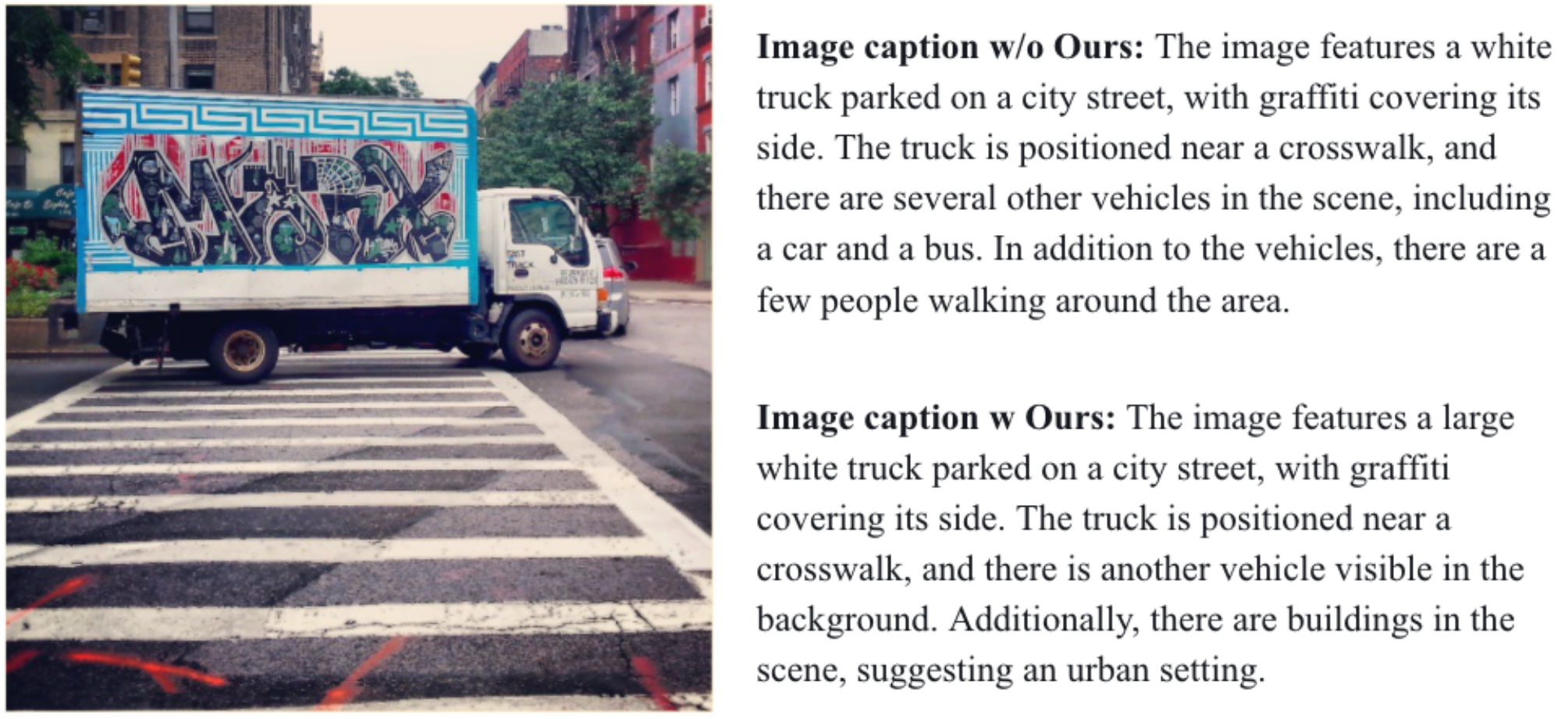}
    \caption{Example of our method applied to an occluded image.}
    \label{fig:occlision}
\end{figure*}
We include a qualitative example in Fig.~\ref{fig:occlision}, where a sedan and a building are partially occluded by a white truck, our method prevents the baseline model from hallucinating of persons and vehicles behind the occluding object. This demonstrates our method remains effective on images consisting of occlusions.
Image caption w/o Ours: The image features a white truck parked on a city street, with graffiti covering its side. The truck is positioned near a crosswalk, and there are several other vehicles in the scene, including a car and a bus. In addition to the vehicles, there are a few people walking around the area. Image caption w Ours: The image features a large white truck parked on a city street, with graffiti covering its side. The truck is positioned near a crosswalk, and there is another vehicle visible in the background. Additionally, there are buildings in the scene, suggesting an urban setting.

\section{Broader Impacts}
\label{sec:impacts}
Our method enhances the factual reliability of multi-modal language models, not only for vision–language tasks but also for modalities such as video and audio, by mitigating hallucinations at inference time. This improvement has several positive societal implications: it can make systems for visual question answering, assistive technologies for the visually impaired, and automated image captioning more dependable, thereby increasing user trust and safety; it can power educational tools that generate accurate descriptions of complex diagrams or historical media, benefiting learners and instructors; and in critical domains such as medical imaging or remote sensing, it can reduce spurious outputs and support more robust decision-making. Conversely, if deployed within surveillance or facial-recognition systems, stronger multi-modal grounding could facilitate more intrusive inferences about individuals from visual data, exacerbating privacy risks.


\end{document}